\documentclass[journal]{IEEEtran}
\usepackage[T1]{fontenc}
\usepackage[inline]{enumitem}
\usepackage[normalem]{ulem} 
\usepackage{times}
\usepackage{helvet}
\usepackage{courier,soulutf8}

\usepackage[utf8]{inputenc}

\usepackage{cite}

\ifCLASSINFOpdf
   \usepackage{graphicx}
   \graphicspath{{images}}
\else
\fi
\usepackage[table]{xcolor}
\usepackage{bm,tikz}
\usepackage{centernot}

\usepackage{url}
\RequirePackage{hyperref}
\PassOptionsToPackage{hyphens}{url}\usepackage{hyperref}
\hypersetup{
	colorlinks=true,
	linkcolor=blue,
	filecolor=magenta, 
	urlcolor=cyan,
	pdftitle={Overleaf Example},
	pdfpagemode=FullScreen,
}
\usepackage{xurl}
\hypersetup{pdfauthor=whatever}

\usepackage{bm}
\def\BibTeX{{\rm B\kern-.05em{\sc i\kern-.025em b}\kern-.08em
    T\kern-.1667em\lower.7ex\hbox{E}\kern-.125emX}}

\usetikzlibrary{matrix}
\usetikzlibrary{arrows,plotmarks,decorations.markings,trees,shapes}
\tikzstyle{n}=[ellipse,draw=black!100,fill=black!10,line width=.7pt,minimum width=1cm,align=center,text height=.2cm]
\tikzstyle{nodo}=[ellipse,draw=black!100,fill=black!30,line width=.7pt,minimum width=1.2cm,text width=2.2cm,align=center,minimum height=.5cm]
\tikzstyle{Qnodo}=[ellipse,draw=black!100,fill=black!10,line width=.7pt,minimum width=1.2cm,,text width=2.2cm,align=center,minimum height=.7cm]
\tikzstyle{arco}=[draw=black!80,line width=1pt, postaction={decorate}, decoration={markings,mark=at position 1.0 with {\arrow[ draw=black!80,line width=.7pt]{>}}}]
\tikzstyle{decision} = [rectangle, draw, fill=black!100,text=white, text width=4.5em, text badly centered, node distance=3cm, minimum height=3em]
\tikzstyle{block} = [rectangle, draw, fill=blue!20, text width=5em, text centered, rounded corners, minimum height=3em]
\tikzstyle{line} = [draw, -latex']
\tikzstyle{cloud} = [draw, ellipse,fill=red!20, node distance=3cm, minimum height=2em]

\def\BibTeX{{\rm B\kern-.05em{\sc i\kern-.025em b}\kern-.08em
    T\kern-.1667em\lower.7ex\hbox{E}\kern-.125emX}}

\usepackage{amssymb}
\usepackage{mathtools} 
\usepackage{amsmath,amsfonts}
\usepackage{textcomp}   
\usepackage{wasysym}
\newcommand{\vempty}{{\centernot{\text{\Circle}}}}

\usepackage[capitalize]{cleveref}

\usepackage{booktabs}
\usepackage{makecell}
\usepackage{multirow}
\usepackage{multicol}
\usepackage{diagbox}
\usepackage{longtable}
\usepackage{tabularx}
\usepackage{array, varwidth, cellspace}

\newcolumntype{L}[1]{>{\raggedright\arraybackslash}S{m{#1}}}
\newcolumntype{M}[1]{>{\centering\arraybackslash}S{m{#1}}}
\newcolumntype{R}[1]{>{\raggedleft\arraybackslash}S{m{#1}}}

\newcommand{\PreserveBackslash}[1]{\let\temp=\\#1\let\\=\temp}
\newcolumntype{C}[1]{>{\PreserveBackslash\centering}p{#1}}

\newcommand\NameEntry[1]{%
  \multirow{3}*{%
    \begin{varwidth}{5em}
    \flushright #1%
    \end{varwidth}}}

\usepackage{url}

\usepackage{easyReview}
\usepackage{highlight}

\newcommand\highlightReference[1]{%
  \expandafter\newcommand\csname highlightReference-#1\endcsname{}%
}
\let\oldbibitem\bibitem
\def\bibitem#1 #2\par{%
  \expandafter\ifx\csname highlightReference-#1\endcsname\relax
    \oldbibitem{#1}#2\par
  \else
    \oldbibitem{#1}\highlightref{#2}\par
  \fi
}
\newcommand\highlightref[1]{\hl{#1}}

\usepackage{subfigure}
 
\begin{document}
%

\title{Rubric-based Learner Modelling via Noisy Gates Bayesian Networks for Computational Thinking Skills Assessment}
%
%
%

\author{Giorgia~Adorni~$^1$,
        Francesca~Mangili~$^1$,
        Alberto~Piatti~$^2$,
        Claudio~Bonesana~$^1$,
        and~Alessandro~Antonucci~$^1$\\
        $^1$ Istituto Dalle Molle di Studi sull'Intelligenza Artificiale (IDSIA) USI - SUPSI, Lugano, Switzerland,\\
        $^2$ Department of Education and Learning (DFA), SUPSI, Lugano, Switzerland
\thanks{This research was funded by the Swiss National Science Foundation (SNSF) under the National Research Program 77 (NRP-77) Digital Transformation (project number 407740\_187246).}}

\maketitle

\begin{abstract}
In modern and personalised education, there is a growing interest in developing learners' competencies and accurately assessing them.
In a previous work, we proposed a procedure for deriving a learner model for automatic skill assessment from a task-specific competence rubric, thus simplifying the implementation of automated assessment tools. 
The previous approach, however, suffered two main limitations: 
(i) the ordering between competencies defined by the assessment rubric was only indirectly modelled; (ii) supplementary skills, not under assessment but necessary for accomplishing the task, were not included in the model. 
In this work, we address issue (i) by introducing dummy observed nodes, strictly enforcing the skills ordering without changing the network's structure. In contrast, for point (ii), we design a network with two layers of gates, one performing disjunctive operations by noisy-OR gates and the other conjunctive operations through logical ANDs.
Such changes improve the model outcomes' coherence and the modelling tool's flexibility without compromising the model's compact parametrisation, interpretability and simple experts' elicitation.
We used this approach to develop a learner model for Computational Thinking (CT) skills assessment. 
The CT-cube skills assessment framework and the Cross Array Task (CAT) are used to exemplify it and demonstrate its feasibility.
\end{abstract}

\begin{IEEEkeywords}
Learner modelling; Bayesian networks with noisy gates; Assessment rubrics; Computational thinking skills.
\end{IEEEkeywords}

%
\IEEEpeerreviewmaketitle

\section{Introduction}
\IEEEPARstart{I}{ntelligent} Tutoring Systems (ITSs) are technological devices that support learning without the mediation of a teacher. They interact directly with the user, providing hints and suggestions that can only be effective if calibrated to the actual user's competence level. 
ITSs collect data on a learner's performance while accomplishing a task and use that data to develop a competence profile based on a predefined model of the learner's knowledge and behaviour. This profile helps determine the most appropriate intervention.
The new knowledge collected along with the learning activity continuously updates the competence profile, making the interventions more focused. 
Therefore, the learner model is one of the main factors that contribute to the success of an AI-based educational tool. 

A learner model describes mathematically the learner's competencies, represented by a set of hidden variables, and their relations with the observable actions performed while solving the task.
Such competencies combine knowledge, skills, and attitudes expressed in a specific context.   
Teachers can evaluate student competencies in realistic scenarios explicitly designed for this purpose and then compare their performance with a model of competence specified through an assessment rubric \cite{dawson2017assessment}. 
A rubric for assessing a student's performance consists of a list of competence components to be evaluated, a qualitative description of possible observable behaviours corresponding to different levels of such components, and a set of criteria for assessing the level of each component. A rubric, therefore, describes the relationship between competencies and observable behaviours of the learner that need to be formally codified. 

Several sources of uncertainty and variability may affect the relationship between the non-observable competencies and the corresponding observable actions. Therefore, a deterministic relationship is not capable of accurately modelling it. Instead, a more appropriate approach would be to use probabilistic reasoning to translate qualitative assessment rubrics into a quantitative, standardised, coherent measure of student proficiency.
In the literature, \emph{Bayesian Knowledge Tracing} (BKT) \cite{corbett1994knowledge}, \emph{Item Response Theory} (IRT) \cite{embretson2013item}, and \emph{Bayesian Networks} (BNs) \cite{koller2009probabilistic} are all popular probabilistic approaches to learner knowledge modelling.
BNs are a robust framework for modelling dependencies between skills and students' behaviours in complex tasks.
In addition, the graphical nature of the models makes them easily understandable by domain experts. Therefore, experts can easily use them in eliciting the student model \cite{millan2000adaptive}. 
Desmarais \cite{desmarais2012review} reviewed all the most successful ITS experiences since Bloom's keynote paper and recognised and presented BNs as the most general approach to modelling learner skills.
\mbox{\cite{mousavinasab2021intelligent}} conducted a systematic review of 53 papers about ITS applications from 2007 to 2017. The review explored the characteristics, applications, and evaluation methods of ITSs and found that a significant proportion of the reviewed papers used BN techniques.
More recent works using BNs to model the learner knowledge in the context of ITSs include 
 \mbox{\cite{hooshyar2018sits}}, who developed an ITS to help students acquire problem-solving skills in computer programming, and \mbox{\cite{xing2021automatic}}, who developed an automatic assessment method for students' engineering design performance using a BN model.
Other recent works, such as \mbox{\cite{wu2020student,rodriguez2021bayesian}}, support the construction ITSs based on BNs and further highlight the ever-present interest in BN techniques for ITSs.
Building on these results, we focused our method on BN-based learner modelling. 

Not all BNs are easy to design, and a deep understanding of BN theory is required. Although BN arcs can be interpreted as a causal model, their definition by experts is not always trivial because of the complexity of the causal relationships involved and the presence of hidden causes.
In addition, a significant effort may be necessary to obtain the network structure and parameters through expert knowledge or the availability of an extensive dataset to learn them directly.

Decomposing domain knowledge into individual basic component skills is the choice typically adopted to simplify the model elicitation process. However, as pointed out in \cite{huang2017learner}, ``complex skill mastery requires not only the acquisition of individual basic component skills but also practice in integrating such component skills with one another''. 
Further complications can also arise even when the structure of the learning model can be precisely defined. In some cases, eliciting and learning BN parameters can quickly make the computation of inferences unmanageable.
The number of parameters and the problem complexity can rapidly increase with the network's number of arcs. 

These issues can discourage ITS practitioners from using BN-based learner modelling in their applications when many skills are involved in the learner's actions. A solution to reduce the number of parameters in a BN-based learner model was proposed in our previous paper \cite{Anonymous2022}. We exploit \emph{noisy-OR gates} \cite{pearl1988probabilistic} to reduce the number of parameters to elicit from exponential in the number of parent skills for each observable action to linear. Similar advantages also concern the inference. In \cite{softcom}, we adopted a solution to set up a general approach for translating assessment rubrics into interpretable BN-based learner models with a complexity compatible with real-time assessment. 

Learner models based on assessment rubrics are more accessible to teachers, who are typically more familiar with them than probabilistic graphical models. Moreover, rubrics focus on the learners' complex behaviours in specific contexts. 
Therefore, although we do not explicitly model skill interactions, they can be captured in the hierarchy of complex behaviours identified by the rubric. 
However, in our previous work \cite{softcom}, this hierarchy of competencies was only indirectly modelled. 
This produced assessments assigning larger probabilities to higher-level competence, in contrast with the assumption that when a competence of a certain level is possessed, all lower-level competencies are also owned. 

In the model presented here, this counter-intuitive behaviour has been eliminated by imposing the constraints codified by the rubric through observed auxiliary nodes that do not modify the network structure and retain the previous model's relative simplicity. Moreover, considering that assessment rubrics are usually limited to the competence components under assessment, hereafter also called \textit{target skills}, the modelling approach in \cite{softcom} may have led to oversimplified learner models, unable to grasp the actual causes of a failure which is not always due to the absence of the target skills. 
It may, for instance, follow from deficiencies in other skills necessary for the specific task, hereafter referred to as \textit{supplementary skills}. 

Therefore, this work extends our previous model \cite{softcom} by adding the possibility of modelling a set of supplementary skills necessary, in conjunction with the skills under examination, to succeed in the  assigned tasks. 
To this goal, it is essential to extend the nosy-OR approach in \cite{Anonymous2022} to combine disjunctive and conjunctive relations between skills since. At the same time, the behaviours in the assessment rubric are mutually exclusive (OR), and supplementary and target skills must be expressed jointly (AND).

To illustrate this approach, we focus on the activity proposed in \cite{piatti_2022} for the standardised assessment of algorithmic skills along the entire K-12 school path. We compare four learner models based on different assumptions and sets of expert-elicited parameters and apply them to the dataset collected in \cite{piatti_2022}. 
Overall, we obtain a general and compact approach to implementing a learner model given a set of competencies of interest and the corresponding assessment rubric. 
The resulting model has a simple structure and interpretable parameters, allowing for fast inferences with a reasonable effort for their elicitation by experts.

This article is organised as follows: 
\begin{itemize}
    \item Section \ref{subsec:bn} provides some background about learner modelling based on Bayesian Networks, and noisy gates.
    \item Section \ref{subsec:rubric} introduces task-specific assessment rubrics.
    \item Section \ref{subsec:rubrictobn} illustrates how to model a generic assessment rubric by Bayesian Networks.
    \item Section \ref{sec:casestudy} presents the case study and, in particular, the specific assessment rubric developed for the Cross Array Task (CAT) from \cite{piatti_2022} as well as the procedure for translating it into a learner model.
    \item Section \ref{sec:results} discusses the model outcomes based on the dataset of 109 pupils collected in  \cite{piatti_2022}. 
    \item Section \ref{sec:concl} summarises the work's findings and contribution. 
\end{itemize}

\section{Method}\label{sec:method}

\subsection{BN-based Learner Models} \label{subsec:bn}
The structure of a Bayesian Network (BN) over a set of variables is described by a directed acyclic graph \begin{math}\mathcal{G}\end{math} whose nodes are in one-to-one correspondence with the variables in the set. We call parents of a variable \begin{math}X\end{math}, according to \begin{math}\mathcal{G}\end{math}, all the variables connected directly with $X$ with an arc pointing to it. Learner models usually include a set of $n$ latent (i.e., hidden) variables \begin{math}\bm{X}:=(X_1,\ldots, X_n)\end{math}, henceforward referred to as \emph{skill nodes}, describing the competence profile of the learner and some $m$ manifest variables \begin{math}\bm{Y}:=(Y_1,\ldots, Y_m)\end{math}, hereafter called \emph{answer nodes}, describing the observable actions implemented by the learner to answer each specific task. 
Arcs go from skill nodes to answer nodes modelling how the presence or absence of a specific competence directly affects the learner's behaviour in a task requiring such competence.
Here, we consider only binary skill nodes, where value one, or \emph{true} state, indicates that the pupil possesses the skill, and binary answers nodes, denoting a correct answer or behaviour shown in solving the task. 

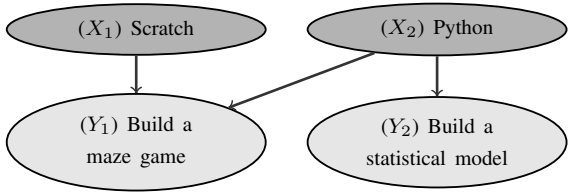
\begin{figure}[htbp]
\centerline{
\begin{tikzpicture}
\node[nodo] (s1)  at (0,0) {\footnotesize (\begin{math}X_1\end{math}) Scratch};
\node[nodo] (s2)  at (4,0) {\footnotesize (\begin{math}X_2\end{math}) Python};
\node[Qnodo] (q1)  at (0.,-1.5) {\footnotesize (\begin{math}Y_1\end{math}) Build~a maze game};
\node[Qnodo] (q2)  at (4,-1.5) {\footnotesize (\begin{math}Y_2\end{math}) Build~a statistical model};
\draw[arco] (s1) -- (q1);
\draw[arco] (s2) -- (q1);
\draw[arco] (s2) -- (q2);
\end{tikzpicture}}
\caption{Example of BN-based learner model.}
\label{fig:bn}
\end{figure}

The relations of a BN-based learner model can be graphically depicted as in the example of Fig.~\ref{fig:bn}. 
The answer nodes describe whether the learner has been able or not to program, for example, a maze game ($Y_1$) or a statistical model ($Y_2$). 
The skill nodes represent the ability to build this program using a block-based programming language such as Scratch ($X_1$) or a text-based programming language such as Python ($X_2$). 
The second skill can be applied to answer both questions, and therefore \begin{math}X_2\end{math} is a parent node for both answer nodes \begin{math}Y_1\end{math} and \begin{math}Y_2\end{math}. 
Instead, the first skill can be used to answer just the first question, and therefore there is no direct arc from \begin{math}X_1\end{math} to \begin{math}Y_2\end{math}. 

Once the graph \begin{math}\mathcal{G}\end{math} structuring the BN is established, the definition of the BN over the $n+m$ variables of the network \begin{math}\bm{V} := (V_1, V_2,\dots, V_{n+m})\end{math}, including both skills ($\bm{X}$) and answers ($\bm{Y}$), consists in a collection of Conditional Probability Tables (CPTs) giving the probabilities
\begin{math}P(Y_i = 1|\mathrm{Pa}(Y_i))\end{math} that \begin{math}Y_i\end{math} takes value one given all possible joint states of its parent nodes \begin{math}\mathrm{Pa}(Y_i)\end{math}. 
Let $\bm{V}$ take values in $\Omega_{\bm{V}}$, the independence relations imposed from \begin{math}\mathcal{G}\end{math} by the \emph{Markov condition}, i.e., the fact that each node is assumed to be independent of its non-descendants non-parents given its parents, induce a joint probability mass function over the BN variables that factorises as follows \cite{koller2009probabilistic}:
\begin{equation}\label{eq:joint}
P(\bm{V}=\bm{v}) = \prod_{v \in \bm{v}} P(v|\mathrm{pa}(V))\,,
\end{equation}
where \begin{math}\bm{v}=(v_1, v_2,\dots, v_{n+m})\end{math} represents a given joint state of the variables in \begin{math}\bm{V}\end{math}.
BN inference consists of the computation of queries based on Eq. \eqref{eq:joint}. In particular, we are interested in \emph{updating} tasks consisting in the computation of the marginal posterior probability mass function for a single skill node \begin{math}X_q \in \bm{X}\end{math} given the observed state $\bm{y}_E$ of the answer nodes $\bm{Y}_E \subseteq \bm{Y}$:
\begin{equation}\label{eq:updating}
P(x_q|\bm{y}_E) = \frac{\sum_{\bm{v}\in \Omega_{\bm{V}|(x_q,\bm{y}_E)}} \prod_{v\in \bm{v}} P(v|\mathrm{pa}(V))}{\sum_{\bm{v}\in \Omega_{\bm{V}|\bm{y}_E}} \prod_{v\in \bm{v}} P(v|\mathrm{pa}(V))}\,,
\end{equation}
where \begin{math}\Omega_{\bm{V}|\bm{v}'}:=\{\bm{v}: v_i = v'_i ~\forall ~v'_i \in \bf{v}'\}\end{math}. 

According to the above model, multiple parent skills may be relevant to the same answer. 
The paper in \cite{Anonymous2022} discusses how this can lead to a critical complexity in elicitation and inference and demonstrates how using noisy gates can avoid these issues.
In this work, we exploit the disjunctive noisy-OR gates, which shape interchangeable skills and are suitable for modelling assessment rubrics.

\subsubsection{Noisy-OR gates}\label{par:noisyor}

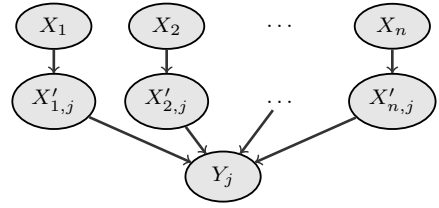
\begin{figure}[htp!]
\footnotesize
\centering
\begin{tikzpicture}
\node[n] (xx1)  at (0,1) {${ X_1 }$};
\node[n] (xx2)  at (1.5,1) {${ X_2}$};
\node[] (xx3)  at (3,1) {${\ldots}$};
\node[n] (xx4)  at (4.5,1) {${ X_n}$};
\node[n] (x1)  at (0,0) {${ X_{1,j}'}$};
\node[n] (x2)  at (1.5,0) {${ X_{2,j}'}$};
\node[] (x3)  at (3,0) {${ \ldots}$};
\node[n] (x4)  at (4.5,0) {${ X_{n,j}'}$};
\node[n] (y)  at (2.25,-1) {${ Y_j}$};
\draw[arco] (xx1) -- (x1);
\draw[arco] (xx2) -- (x2);
\draw[arco] (xx4) -- (x4);
\draw[arco] (x1) -- (y);
\draw[arco] (x2) -- (y);
\draw[arco] (x3) -- (y);
\draw[arco] (x4) -- (y);
\end{tikzpicture}
\caption{A noisy gate explicit formulation (adapted form \cite{Anonymous2022}).}
\label{fig:bn2}
\end{figure}

This section briefly introduces noisy-OR gates and their use in learner modelling. We refer to \cite{Anonymous2022} and \cite{softcom} for a more detailed discussion. 
A typical representation of the noisy-OR network structure, introducing \begin{math}n\end{math} auxiliary variables (also called inhibitor nodes), is shown in Fig.~\ref{fig:bn2}. 

To reduce the number of parameters, the noisy-OR defines the state of \begin{math}Y_j\end{math} as the logical disjunction (OR) of the auxiliary parent nodes, removing the need to specify the CPT of the answer node given the state of its parent nodes. Furthermore, the noisy-OR structure sets the input variable \begin{math}X_i\end{math} as the unique parent of \begin{math}X_{i,j}'\end{math} and constraints \begin{math}X'_{i,j}\end{math} to be zero with probability one when \begin{math}X_i=0\end{math}.
The relationship between skill and answers would be purely logical-deterministic were it not for the noise introduced by the so-called inhibition parameters $\lambda_{i,j} = P(X_{i,j}'=0|X_i=1)$, representing the probability of not expressing skill $i$ in task $j$.
Auxiliary variables can be interpreted as \emph{inhibitors} of the corresponding skills, as they can be in the false state, e.g., \begin{math}X'_{i,j} = 0\end{math} (with probability \begin{math}\lambda_{i,j}\end{math}) even when the corresponding skill node $X_i$ is in the true state, indicating that, although possessed by the learner, the skill could not be expressed in task \begin{math}Y_j\end{math}. By defining the probability of a failure in expressing a possessed skill in the specific task $j$, the inhibition parameter $\lambda_{i, j}$ provides a measure of the task difficulty.  
If a pair skill-answer has a large inhibition, the state of the answer node tells, in general, little about the state of the skill node, the extreme case of $\lambda_{i,j} = 1$ corresponding to a missing arc in the BN graph between skill $i$ and answer $j$.

The noisy-OR network induces the following CPT between the $n$ parent skill nodes $\bf{X} = (X_1,\dots,X_n)$ and the observable answer node $Y_j$ \cite{pearl1988probabilistic}:
\begin{equation}\label{eq:noisy}
P(Y_j=0|\bf{X} = (x_1,\ldots,x_n)) = \prod_{i=1}^n (\mathbb{I}_{x_i=0}+\lambda_i\mathbb{I}_{x_i=1})\,,
\end{equation}
where \begin{math}\mathbb{I}_A\end{math} is the indicator function returning one if \begin{math}A\end{math} is true and zero otherwise. 
The second term $\lambda_i\mathbb{I}_{x_i=1}$ represents the noise as it introduces the possibility that a skill $X_i$ that the student possesses is not expressed in task $Y_j$ (this phenomenon is also called \textit{slip} elsewhere in this work). The value of $\lambda_i$ implying the biggest uncertainty associated with the task-skill pair ($Y_j$, $X_i$)  is 0.5, whereas the value $\lambda_i = 0$ models the certainty that skill $X_i$, whenever present, will be expressed in solving task $Y_i$ and, vice versa, $\lambda_i = 1$  model the fact that $X_i$ cannot be expressed in task $Y_j$.

In \cite{softcom}, a leak node was added to the model to represent the possibility of a random guess, i.e., a correct answer or a behaviour given without mastering any required competencies. The leak is a Boolean variable playing the role of an additional skill node, \begin{math}X_{\mathrm{leak}}\end{math}, which is set in the observed state $X_{\mathrm{leak}}=1$ and added to the parents of all answer nodes for which random guessing is possible. The chances of guessing answer $Y_j$ at random are given by parameter $1-\lambda_{leak,j}$. 

To apply the above model, the domain expert (e.g., the teacher) should first list the parentless skill nodes (including, eventually, the leak) \begin{math}X_1, \dots, X_n\end{math}, the childless answer nodes \begin{math}Y_1, \dots, Y_m\end{math} and connect by an arc the skills to all answer nodes in which they can be used. 
Then, the instructor should quantify for each pair of skill-answer nodes, $X_i$ and $Y_j$, connected by an arc, the value of the inhibition \begin{math}\lambda_{i,j}\end{math}. This results in a total of at most $n\cdot m$ parameters to be elicited. Finally, the expert should state each skill's prior probabilities $\pi_i$.

\subsubsection{Comparison with BKT} 
While the BKT, in its standard implementation, traces the evolution of a single skill over time, our approach focuses on fine-grained skills modelling at the specific moment the assessment is performed. However, a parallel can be drawn between the two. BKT models student knowledge at time $t$ as the (binary) latent variable $X(t)$ of a hidden Markov model \cite{corbett1994knowledge}. Learning is modelled as the transition of $X(t)$ from state zero (lack of knowledge) to state one (knowledge acquired). The model defines four parameters: (i) the \textit{initial} probability, i.e., the probability that the knowledge has been already acquired at the beginning of the activity; (ii) the \textit{learning} probability, that is, the probability of acquiring the probability between $t$ and $t+1$; (iii) the \textit{slip} probability of making a mistake when the knowledge is acquired; (iv) the \textit{guess} probability of doing right in the lack of knowledge. 

In our model, the probability of the \textit{slip} may vary depending on the pair skill $i$ and task $j$ and is represented by the inhibition $\lambda_{ij}$. The  \textit{guess} probability depends on the task and is equal to $1-\lambda_{\mathrm{leak},j}$. The \textit{initial} probability of a skill $X_i$ is defined by its prior probability $\pi_i$. 
Notice, however, that since our approach, differently from BKT, does not model the learning process, the concept of initial probability here is meant to describe our initial knowledge of the learner competence profile rather than the probability that the skill is initially acquired. For the same reason, no \textit{learning} probability is defined in our model. 

 \subsection{Assessment Rubrics}\label{subsec:rubric}

There are several possible approaches to identifying the knowledge components to be included in a learner model. In this article, we follow the one introduced in \cite{softcom} and start from a rubric defined for assessing a given competence through a specific task or family of similar tasks \cite{castoldi2009valutare,jonsson2007use}.

A task-specific assessment rubric consists of a two-entry table where each row corresponds to a component of the given competence, described in the light of the given task. In contrast, each column corresponds to a competence level in ascending order of proficiency. 
For each combination of component and level, the rubric provides a qualitative description of the behaviour expected from a person with the given level in the given component.
Identifying a person's competence level consists in matching the learner's behaviours while solving a given task with those described in the assessment rubric. 

\begin{table}[!htb]
\begin{center}
\caption{Example of a task-specific assessment rubric with a single competence component, the ability to design algorithm containing loops, and two competence levels.\label{tab:examplerubric}}
\begin{tabular}{p{.6cm}|p{3.5cm}p{3.5cm}}
&\multicolumn{1}{c}{{$X_1$}}    & \multicolumn{1}{c}{{$X_2$}}   \\
\midrule
\multirow{3}{*}{Loops} & Develop an iterative algorithm using a block-based programming language & Develop an iterative algorithm using a text-based programming language     \\
\end{tabular}
\end{center}
\end{table}
For instance, Table~\ref{tab:examplerubric} shows the task-specific assessment rubric for an example focused on assessing the student's ability to use iterative instructions in algorithms. 
This competence has two levels depending on the tools used by the learner: a visual programming language ($X_1$) or a textual programming language ($X_2$).  
By checking how the learner produced the algorithm, the teacher can see whether he applied any of the methods in the rubrics and assign him the corresponding competence level.

Generally speaking, assessment rubrics define an ordering between competence levels and sometimes between competence components, as for the case study about computational thinking introduced in \cite{softcom} and discussed in Section \ref{sec:casestudy}.  
Here, a competence level or component is considered higher than another if the former implies the latter, meaning that a learner with the higher competence can also perform all the tasks that require the lower. 
In practice, the competence level matching the student's behaviours for a given component does not always correspond to the actual learner's state of knowledge. It is also possible that the person possesses a higher level but is underperforming.

In the case of a task composed of similar sub-tasks, i.e., tasks sharing the same assessment rubric, it is, therefore, possible to observe behaviours matching different competence levels in the various sub-tasks. 
In the following subsection, we illustrate how this uncertainty can be considered and how an overall assessment based on a full battery of tasks can be produced by modelling the learner competence profile with the BN-based approach described in Section \ref{subsec:bn}. 

\subsection{Modelling Assessment Rubrics by Bayesian Networks}\label{subsec:rubrictobn}
Considering a task-specific assessment rubric, as defined in Section \ref{subsec:rubric}, it is possible to derive a learner model, as presented in Section \ref{subsec:bn}, hereafter referred to as \emph{baseline model}.
For each cell $(c,r)$ of an assessment rubric with \begin{math}R\end{math} rows and \begin{math}C\end{math} columns, we introduce a latent binary competence variable $X_{rc}$ taking value one for a learner mastering the corresponding competence level and zero otherwise. 
Moreover, for each task $t$, in a battery of $T$ similar tasks, and each competence variable $X_{rc}$, we define an observable (manifest) binary variable \begin{math}Y^t_{rc}\end{math} taking value one if the behaviour described in the assessment rubric's cell $(r,c)$ was applied successfully by the learner in solving task $t$ and zero if he failed using it. 

In this work, we improve the baseline model in two ways. Firstly, we explicitly impose the ordering of competence levels encoded by the rubric. Secondly, we include in the model task-specific supplementary skills which can be combined with each other and with the competencies of the rubric through arbitrary logic functions.

\subsubsection{Ordering of competences} \label{par:dummies}
In the baseline model of Section \ref{subsec:bn}, it was indirectly accounted for the partial ordering between variables by setting as parents of answer node $Y^t_{rc}$ the skill node $X_{rc}$ and all skill nodes corresponding to higher competence levels. The network was quantified through noisy-OR relations, as described in Section \ref{par:noisyor}.
This structure assumes that an observed behaviour can be explained as the student mastering the corresponding competence level or a higher one if he is underperforming, thus not exploiting his full potential, but cannot be achieved through a lower level. 

As mentioned above, we interpret the (partial) ordering between competencies defined by the assessment rubric as implication constraints, meaning that possessing a particular skill $X_i$ implies that the learner posses also his inferior competencies. While exploited to design the network structure, this hierarchy of competencies is not strictly imposed by the above baseline model, giving rise to posterior inferences that are usually inconsistent. 

To solve this issue, we enrich the model by adding an auxiliary variable $D_{ik}$ for each relation $X_i \implies X_k$ defined by the rubric. 
A constraint node $D_{ik}$ is always in the observed state one and has $X_i$ and $X_k$ as parent nodes. The desired implication constraint is then implemented by choosing a CPT for $D_{ik}$ such that $P(D_{ik}=1|X_i=1, X_k=0) = 0$. 
The addition to the network of each observed node $D_{ik}$ changes the prior probabilities of $X_i$ and $X_k$, initially set to $\pi_i$ and $\pi_k$. 
Let 
\begin{equation}
\begin{split}
p_{00} & = P(D_{ik}=1|X_i=0, X_k=0) \\
p_{01} & = P(D_{ik}=1|X_i=0, X_k=1) \\
p_{11} & = P(D_{ik}=1|X_i=1, X_k=1),
\end{split}
\end{equation}
be the non-null parameters in the CPT of $D_{ik}$.
After updating with the evidence $D_{ik}=1$, one has 
\begin{equation}\label{eq:priorwithD}
\begin{split}
&P(X_i=1|D_{ik}=1) = \frac{p_{11}\pi_j \pi_k}{K},\\
&P(X_k=1|D_{ik}=1) = \frac{p_{11}+p_{01}\pi_j \pi_k}{K},
\end{split}
\end{equation}
with $K= p_{11}\pi_j \pi_k+p_{01}(1-\pi_j) \pi_k+p_{00}(1-\pi_j) (1-\pi_k )$.

In this work, we simply assume $p_{00} = p_{01} =p_{11}$ and adopt uniform prior probabilities $\pi_i=\pi_k=0.5$. 
Applying them to Eq.(\ref{eq:priorwithD}) give $P(X_k=1) = 1/3$ and $P(X_k=1) = 2/3$. 
This result follows from the fact that skill $X_i$ can only be possessed jointly with $X_k$, whereas $X_k$ can also be owned when $X_i=0$. 

Under the assumption $p_{00} = p_{01} =p_{11} = p_{*}$, the prior over the superior skill $X_i$ can be interpreted as the conditional probability of having it given that the learner possesses the inferior skill $X_k$ since 
\begin{equation}
\begin{split}
P(X_i=1|X_k=1, D_{ik}=1)  = &\frac{\pi_i\pi_j p_{*}}{\pi_i\pi_j p_{*} + (1-\pi_i)\pi_j p_{*}} = \pi_i
\end{split}
\end{equation}

\subsubsection{Supplementary skills} \label{par:supp}
While the assessment rubric details the components of the competence of interest and their interactions with the specific task and available tools, it does not necessarily include all the skills required to solve the task successfully.

For instance, considering the assessment rubric proposed in Table~\ref{tab:examplerubric}, to develop an iterative algorithm with a text-based programming language successfully, the learner might also need knowledge about the different types of statements, e.g., while, repeat, for, do until and so on.
Ignoring such supplementary skills might be misleading in an automatic assessment system, as failures due to the lack of one of them would not be recognised as such and, eventually, be attributed to the absence of the competence components under assessment. Therefore, if not adequately modelled, the lack of unmodelled supplementary skills would translate into an unfairly negative evaluation of the competencies of interest. 

To produce fairer assessments, we extend the model by an additional layer of auxiliary nodes combined with a logic function to allow for the inclusion of a suitable set of supplementary skills.

Fig.~\ref{fig:finalbn} shows an example of the structure of the extended network. 
Supplementary skills are described by additional skill nodes $S_1, \dots, S_m$, which are grouped into sets of interchangeable skills (in the case of the example we have just one set). 
Each of these groups is connected through a noisy-OR to a node in the layer of auxiliary latent nodes, hereafter referred to as group nodes $G_1, \dots, G_l$, representing the success or failure in applying the type of competence described by each group to the specific task $Y$. 
Finally, the group nodes are connected to the answer node through a logic AND or any other logic function suitable for the particular task. 

\begin{figure}[htb]
 \centering
 \includegraphics[width=.8\columnwidth]{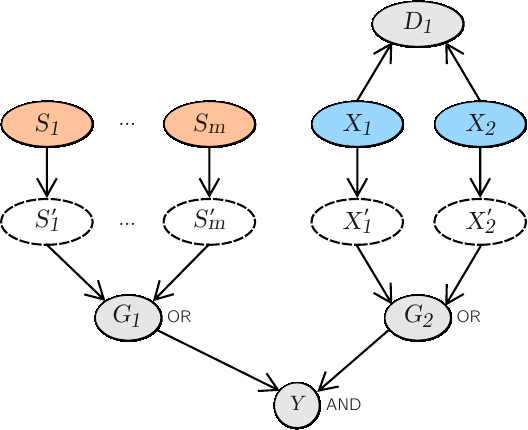}
 \caption{
 Example of a BN network modelling a task-specific assessment rubric with two cells, represented by skills $X_1$ and $X_2$ (on the right), $m$ supplementary skills grouped in a single set (on the left), and the constraint $X_2\implies X_1$, represented by the auxiliary variable $D_1$ (on the top right).}
 \label{fig:finalbn}
\end{figure}

When supplementary skills can be directly assessed through observing specific learner behaviours or by purposed questions, additional answer nodes, children only of the relevant supplementary skills, can be added to the network.  

\section{A Case study on K-12 Computational Thinking skills}\label{sec:casestudy}

In this section, we use the case study about Computational Thinking (CT) skill assessment already introduced in \cite{softcom} to illustrate the proposed approach. However, our methodology can be applied analogously to each task for which an assessment rubric can be defined.

CT assessment is an important field of research \cite{poulakis2021computational} due to its relevance in evaluating the effectiveness of CT teaching and learning activities on an individual or class level and, on a larger scale, the impact of curricular and educational system policies on the development of CT skills of a given population.

\subsection{The Cross Array Task}
The Cross Array Task (CAT), proposed in \cite{piatti_2022}, is an unplugged activity designed to assess the development of algorithmic skills, i.e., the ability to describe a complex procedure through a set of simpler instructions, in pupils aged from 3 to 16 years.

The authors of \cite{piatti_2022} carried out an experimental study from March to April 2021, collecting data from 109 students (51 girls and 58 boys) in eight classes from three public schools in Southern Switzerland. 

\begin{figure}[htb]
 \centering
 \includegraphics[width=.8\columnwidth]{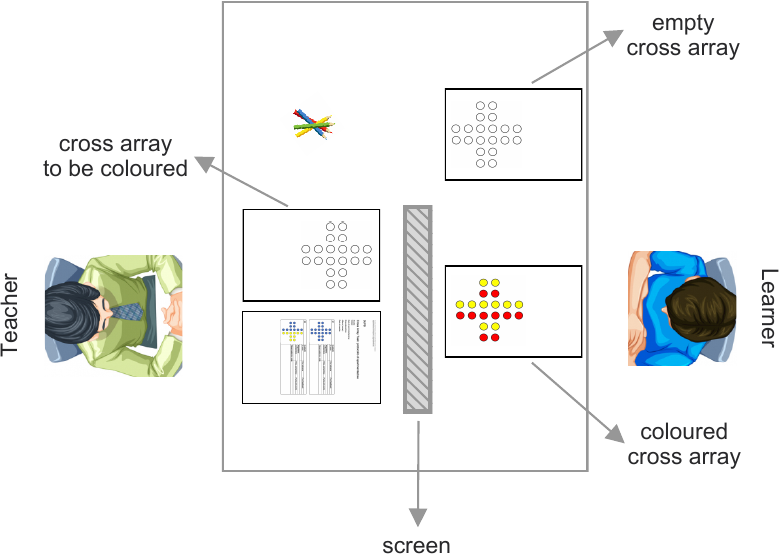}
 \caption{CAT experimental settings (adapted from \cite{piatti_2022}).}
 \label{fig:cat}
\end{figure}

The data collection setting is illustrated in Fig.~\ref{fig:cat}.
In the activity, each student was provided with 12 coloured cross arrays, visible in the right part of Fig.~\ref{fig:cat} and at the top of Fig.~\ref{fig:its}, characterised by different levels of complexity and based on different types of regularities, and requested to give the teacher instructions, either by voice (symbolic artefact) or with the help of an empty cross array schema (embodied artefact), to reproduce the same colouring patterns on a blank cross array.
The pupil and the teacher are initially separated by a screen that does not allow the student to see what the other is doing. Upon request by the pupil, this barrier is removed to allow him/her to see how his commands are implemented (visual feedback).

\subsection{Modelling the CAT assessment rubric}
The assessment rubric for this case study, shown in Table~\ref{tab:catrubric}, is the same as presented in our previous work \cite{softcom}.

The instruction sequences built by the pupils, called algorithms, are ranked into three categories corresponding to the assessment rubric's competence components (rows). Each row represents the ability of the pupil to solve a CAT schema using, respectively, \textit{zero-dimensional (0D)} algorithms – where the dots of the schema are described point by point –, \textit{one-dimensional (1D)} algorithms – where structures, such as rows, diagonal, squares etc. are also used to illustrate the coloured pattern –, and \textit{two-dimensional (2D)} algorithms – where repetitions and loops on dots or structures are also used to describe the schema.

The degree of autonomy of the pupil and the tools used to accomplish the task have been hierarchically ordered and determine the competence levels in the columns of the rubric. Specifically, from the highest (right) to lowest (left), such levels correspond to the ability of the pupil to solve a CAT schema using: \textit{voice (V)} – the pupil can communicate all the necessary instructions using only the voice –, \textit{voice \& schema (VS)} – the pupil, in combination with the voice, uses an empty cross array schema to illustrate his instructions through gestures, e.g., by pointing with fingers the dots to be coloured –, \textit{voice, schema \& feedback (VSF)} – the student also asks to remove the screen to have visual feedback of how the teacher is colouring the schema.

As introduced in Section \ref{subsec:rubrictobn}, the columns of an assessment rubric provide the competence levels in increasing order from left to right. Sometimes, as in this case study, this is true also for the rows, where competence components are ordered from the lower (0D at the top) to the highest (2D at the bottom).  
This follows from the assumption that mastering algorithms of higher complexity imply also mastering simpler ones. The same is valid for communication tools.

Summing up, we can conclude that a competence level $X_{rc}$ is higher than $X_{r'c'}$ whenever $c>c'$ and $r \geq r'$, or $c=c'$ and $r>r'$. When, instead, $c>c'$ but $r<r'$, neither skill can be said to dominate the other. 

From the CAT assessment rubric in Table~\ref{tab:catrubric} we define nine target skills to be assessed, representing the ability to develop an algorithm using elementary operations communicated either by a symbolic language ($X_{13}$ – 0D V), an embodied language ($X_{12}$ – 0D-VS) or supported by visual feedback ($X_{11}$ – 0D-VSF); 
 elementary operations and structures communicated either by a symbolic language ($X_{23}$ – 1D-V), an embodied language ($X_{23}$ – 1D-V) or supported by visual feedback ($X_{21}$ – 1D-VSF);
 elementary operations, structures and repetitions communicated either by a symbolic language ($X_{33}$ – 2D-V), an embodied language ($X_{32}$ – 2D-VS), or supported by visual feedback ($X_{31}$ – 2D-VSF).

\begin{table}[htb]
\begin{center}
\caption{Definition of the CAT assessment rubric.\label{tab:catrubric}}
\begin{tabular}{p{1.5cm}p{1.5cm}|C{.8cm}C{.8cm}C{.8cm}}
\multicolumn{2}{c|}{}  & \multicolumn{3}{c}{{Competence levels}} \\
\multicolumn{2}{c|}{}  & {VSF $c=1$}    & {VS $c=2$}   & {V $c=3$}   \\
\midrule
\NameEntry{{Competence \newline components}} & {0D ($r=1$)} & \begin{math}X_{11}\end{math}   & \begin{math}X_{12}\end{math}  & \begin{math}X_{13}\end{math}             \\
                                             & {1D ($r=2$)}     & \begin{math}X_{21}\end{math}   & \begin{math}X_{22}\end{math}  & \begin{math}X_{23}\end{math}             \\
                                             & {2D ($r=3$)}          & \begin{math}X_{31}\end{math}   & \begin{math}X_{32}\end{math}  & \begin{math}X_{33}\end{math}             \\
\end{tabular}
\end{center}
\end{table}

Accordingly, with the method described in Section \ref{subsec:rubrictobn}, a latent skill node $X_{rc}$ is included in the BN learner model for each of the nine target skills of the rubric. 
The hierarchy of competencies is then modelled by nine latent binary variables $D_{rc,r'c'}$, as described in Section \ref{subsec:rubric}, encoding the implication $X_{rc} \implies X_{r'c'}$ for each pair of consecutive skills in the hierarchy, i.e., such that $(r = r'+1) \land (c= c')$ or $(r = r') \land (c= c'+1)$.

Also, the BN network includes an observable answer node $Y^t_{rc}$ for each skill in the rubric and each task $t = 1, \dots, 12$ in the sequence of 12 similar tasks administered during the CAT experiments. 
Observing \begin{math}Y^t_{rc} = 1\end{math} means that the pupil has solved the $t$-th CAT schema using an algorithm of complexity corresponding to the $c$-th row of the rubric, and requesting the help in the $r$-th column. 
By way of example, a student solving the $t$-th schema using a 0-dimensional algorithm and with voice, empty schema and feedback (0D-VSF) result in the observed node $Y^t_{11}=1$.

In principle, all answer nodes should be explicitly observed through specific interactions with the pupil. However, this was not possible for the case of the dataset collected by \cite{piatti_2022} since the pupils were left free to choose their preferred approach.
Therefore, to make such a dataset compatible with our model, in \cite{softcom} we encode the collected answers as follows: a task $t$ solved at level $c^*$ by an algorithm with complexity $r^*$ was translated into $Y^t_{rc} = 1$ for all competence levels $rc$ lower than or equal to $r^*c^*$, thus assuming that, if requested, the pupil would have been able to implement solutions requiring a lower competence level than the one used. Similarly, we set all answer nodes $Y^t_{rc} = 0$ for all higher levels, leaving those not directly comparable unobserved.
As an example, Table~\ref{tab:1dvs} (i) illustrates the case in which the pupil has generated as a solution for task $t$ a one-dimensional algorithm using only the empty schema and the voice (1D-VS). 

The choice made in our previous work \cite{softcom} also contributed to stress the skills ordering. Since the ordering is modelled by explicit constraints imposed through the auxiliary variables $D_{rc, r'c'}$, such a choice would be unnecessary and detrimental, as it would artificially multiply the number of observations. 
Therefore, in the constrained model, a task $t$ solved at level $c^*$ by an algorithm with complexity $r^*$ would be better translated into the single observation $Y^t_{r^*c^*} = 1$. However, in the specific experimental setting adopted in \cite{piatti_2022}, since pupils were always allowed to try solving the task with the lowest competence level (0D-VSF), a failure could only be observed for that level, with the consequence that only answer nodes $Y^t_{11}$ can be directly observed in the false state $Y^t_{11}=0$. To work around this problem, we set the answer nodes just above the one observed in the true state, i.e., $Y^t_{r^*(c^*+1)}$ and $Y^t_{(r^*+1)c^*}$ to the false state, leaving all other nodes unobserved. 
As an example, Table~\ref{tab:1dvs} (ii) shows how the answer encoding changes in the case of a 1D-VS solution to task $t$ for the constrained model.

\begin{table}[htb]
\setlength\extrarowheight{2pt}
\begin{center}
\caption{Answer encoding, assuming a pupil has generated a 1D-VS solution for the $t$-th schema: \begin{math}Y^t_{22} = 1\end{math}. Symbol $\vempty$ 
 indicates that the answer node is not observed.\label{tab:1dvs}}
\begin{tabular}{p{1.5cm}p{1.5cm}|C{.8cm}C{.8cm}C{.8cm}}
\multicolumn{5}{c}{(i) Unconstrained BN-based learner model}\\
\multicolumn{2}{c|}{\multirow{3}{*}{{$Y^t_{rc}$}}}  & \multicolumn{3}{c}{{Competence levels}} \\
\multicolumn{2}{c|}{}        & {VSF $c=1$}    & {VS $c=2$}   & {V $c=3$}   \\
\midrule
\NameEntry{{Competence \newline components}} & {0D ($r=1$)} & \cellcolor[HTML]{D4EFDF}{1}         & \cellcolor[HTML]{D4EFDF}{1}   & \begin{math}\vempty \end{math}             \\
                                                & {1D ($r=2$)}     & \cellcolor[HTML]{D4EFDF}{1}          & \cellcolor[HTML]{85C1E9}{\textbf{1}}   & \cellcolor[HTML]{FADBD8}{0}            \\
                                                & {2D ($r=3$)}          & \begin{math}\vempty\end{math}   & \cellcolor[HTML]{FADBD8}{0}  & \cellcolor[HTML]{FADBD8}{0}              \\
\end{tabular}
\bigskip
\begin{tabular}{p{1.5cm}p{1.5cm}|C{.8cm}C{.8cm}C{.8cm}}
\multicolumn{5}{c}{(ii) Constrained BN-based learner model}\\
\multicolumn{2}{c|}{\multirow{3}{*}{{$Y^t_{rc}$}}}  & \multicolumn{3}{c}{{Competence levels}} \\
\multicolumn{2}{c|}{}        & {VSF $c=1$}    & {VS $c=2$}   & {V $c=3$}   \\
\midrule
\NameEntry{{Competence \newline components}} & {0D ($r=1$)} & \begin{math}\vempty\end{math} & \begin{math}\vempty\end{math}   & \begin{math}\vempty \end{math}             \\
                                        & {1D ($r=2$)}  & \begin{math}\vempty\end{math}  & \cellcolor[HTML]{85C1E9}{\textbf{1}}  & \cellcolor[HTML]{FADBD8}{0}            \\
                                        & {2D ($r=3$)}  & \begin{math}\vempty\end{math}  & \cellcolor[HTML]{FADBD8}{0}   & \begin{math}\vempty\end{math}              \\
\end{tabular}
\end{center}
\end{table}

Finally, having observed that, depending on the specific CAT schemes, other skills than those in the assessment rubric may be necessary, especially for the algorithm complexities 1D and 2D, by analysing the CAT schemes structures and characteristics, we identify ten supplementary skills, divided into three groups, to be added to the skill nodes in the network:  
($S_1$) {paint dot – group 1}; 
($S_2$) {fill empty dots – group 2}; 
($S_3$) {paint monochromatic rows or columns – group 2}; 
($S_4$) {paint monochromatic squares – group 2}; 
($S_5$) {paint monochromatic diagonals – group 2}; 
($S_6$) {paint monochromatic l-shaped patterns – group 2}; 
($S_7$) {paint monochromatic zigzags – group 2}; 
($S_8$) {paint polychromatic rows or columns – group 3}; 
($S_9$) {paint polychromatic diagonals or zigzags – group 3}; 
($S_{10}$) {repetition of a pattern – group 3}.

Note that, the first group of supplementary skills is made of the variable $S_1$ alone, which is necessary to implement a 0D algorithm. 
The second group includes all the monochromatic structures and is associated with 1D algorithms. 
The last group, which combines all polychromatic structures and the ability to repeat a structure, identifies the skills without which it would not be possible to develop a 2D algorithm. 

From the annotations collected during the experimental study in \cite{piatti_2022} it was possible to extract direct observations about using each supplementary skill in each task. Consequently, answer nodes $Y^t_{S_i}$ were added to the network for each task $t = 1, \dots, 12$ and each supplementary skill $S_i$ with $i=1, \dots, 10$.
Each schema can be solved using one or more supplementary skills, but using all of them is not always possible.
Answer nodes \begin{math}Y^t_{S_i}\end{math} take the value one if the pupil has used the $i$-th supplementary skill in the solution of CAT schema $t$, and zero otherwise.

As described in Section \ref{subsec:rubrictobn}, a noisy-OR combines the variables in the same group into the group auxiliary nodes $G_i$, with $i=1,\dots, 4$, where $G_4$ combines the target skills $X_{rc}$. 
In contrast, the relation between the group nodes and the target skills is conveyed through the logical AND.

\subsection{Summary score metric}

To evaluate pupils' competence level in a specific CAT schema, we use the CAT score, a metric provided by \cite{piatti_2022}, taking values from 0 to 4, as shown in Table~\ref{tab:catscore}. 
Unsolved CAT schemes, which were not considered in \cite{piatti_2022}, are here assigned a score of $-1$. 
\begin{table}[htb]
\caption{The CAT score metric used in \cite{ piatti_2022} to grade a pupil on a single task (adapted from \cite{softcom}). 
\label{tab:catscore}}
\begin{center}
\begin{tabular}{l|ccc}
 & {VSF} & {VS} & {V} \\
\midrule
{0D} & 0 & 1 & 2 \\
{1D} & 1 & 2 & 3 \\
{2D} & 2 & 3 & 4 \\
\end{tabular}
\end{center}
\end{table}

When using the BN-based learner model, a summary metric of the inferences at the end of the 12 tasks, hereafter referred to as the BN-based CAT score, was obtained as the sum of the marginal posterior probabilities of all target skill nodes estimated by the model and can be interpreted as the expected number of competence levels mastered by the student. 

\begin{figure*}[!ht]
\centering
\begin{tikzpicture}[]
\tikzset{b/.style={draw,circle,scale=0.32,blue!40!gray,fill}}
\tikzset{g/.style={draw,circle,scale=0.32,yellow,fill}}
\tikzset{r/.style={draw,circle,scale=0.32,red!80!gray,fill}}
\tikzset{v/.style={draw,circle,scale=0.32,green!40!gray,fill}}
\tikzset{l10/.style={draw,rectangle,scale=0.42,black!99,fill}}   
\tikzset{l9/.style={draw,rectangle,scale=0.42,black!90,fill}}    
\tikzset{l8/.style={draw,rectangle,scale=0.42,black!81,fill}}    
\tikzset{l7/.style={draw,rectangle,scale=0.42,black!72,fill}}    
\tikzset{l6/.style={draw,rectangle,scale=0.42,black!63,fill}}    
\tikzset{l5/.style={draw,rectangle,scale=0.42,black!54,fill}}    
\tikzset{l4/.style={draw,rectangle,scale=0.42,black!45,fill}}    
\tikzset{l3/.style={draw,rectangle,scale=0.42,black!36,fill}}    
\tikzset{l2/.style={draw,rectangle,scale=0.42,black!27,fill}}    
\tikzset{l1/.style={draw,rectangle,scale=0.42,black!18,fill}}    
\tikzset{l0/.style={draw,rectangle,scale=0.42,black!9,fill}}     
\tikzset{n/.style={draw,rectangle,scale=0.42,black!0,fill}}
\matrix[matrix of nodes,row sep=0.1mm,column sep=0.1mm]
{&&&&&|[b]|&|[b]|&&&&&
 &&&&&|[b]|&|[g]|&&&&&
 &&&&&|[g]|&|[r]|&&&&&
 &&&&&|[b]|&|[g]|&&&&&
 &&&&&|[g]|&|[b]|&&&&&
 &&&&&|[b]|&|[v]|&&&&&
 &&&&&|[g]|&|[r]|&&&&&
 &&&&&|[g]|&|[r]|&&&&&
 &&&&&|[b]|&|[v]|&&&&&
 &&&&&|[b]|&|[b]|&&&&&
 &&&&&|[b]|&|[b]|&&&&&
 &&&&&|[b]|&|[v]|&&&&\\
 &&&&&|[b]|&|[b]|&&&&&
 &&&&&|[b]|&|[g]|&&&&&
 &&&&&|[g]|&|[r]|&&&&&
 &&&&&|[b]|&|[g]|&&&&&
 &&&&&|[g]|&|[b]|&&&&&
 &&&&&|[b]|&|[v]|&&&&&
 &&&&&|[r]|&|[g]|&&&&&
 &&&&&|[r]|&|[b]|&&&&&
 &&&&&|[r]|&|[g]|&&&&&
 &&&&&|[g]|&|[g]|&&&&&
 &&&&&|[v]|&|[v]|&&&&&
 &&&&&|[r]|&|[g]|&&&&\\
 &&&|[b]|&|[b]|&|[b]|&|[b]|&|[b]|&|[b]|&&&
 &&&|[b]|&|[b]|&|[b]|&|[g]|&|[g]|&|[g]|&&&
 &&&|[g]|&|[r]|&|[g]|&|[r]|&|[g]|&|[r]|&&&
 &&&|[g]|&|[r]|&|[b]|&|[g]|&|[r]|&|[b]|&&&
 &&&|[r]|&|[r]|&|[g]|&|[b]|&|[v]|&|[v]|&&&
 &&&|[b]|&|[b]|&|[b]|&|[v]|&|[v]|&|[v]|&&&
 &&&|[g]|&|[r]|&|[g]|&|[r]|&|[g]|&|[r]|&&&
 &&&|[g]|&|[r]|&|[g]|&|[r]|&|[g]|&|[r]|&&&
 &&&|[b]|&|[v]|&|[b]|&|[v]|&|[b]|&|[v]|&&&
 &&&|[g]|&|[b]|&|[g]|&|[g]|&|[b]|&|[g]|&&&
 &&&|[v]|&|[b]|&|[r]|&|[r]|&|[b]|&|[v]|&&&
 &&&|[g]|&|[v]|&|[b]|&|[v]|&|[b]|&|[r]|&&\\
 &&&|[b]|&|[b]|&|[b]|&|[b]|&|[b]|&|[b]|&&&
 &&&|[b]|&|[b]|&|[b]|&|[g]|&|[g]|&|[g]|&&&
 &&&|[g]|&|[r]|&|[g]|&|[r]|&|[g]|&|[r]|&&&
 &&&|[g]|&|[r]|&|[b]|&|[g]|&|[r]|&|[b]|&&&
 &&&|[r]|&|[r]|&|[g]|&|[b]|&|[v]|&|[v]|&&&
 &&&|[g]|&|[g]|&|[g]|&|[r]|&|[r]|&|[r]|&&&
 &&&|[r]|&|[g]|&|[r]|&|[g]|&|[r]|&|[g]|&&&
 &&&|[r]|&|[b]|&|[r]|&|[b]|&|[r]|&|[b]|&&&
 &&&|[r]|&|[g]|&|[r]|&|[g]|&|[r]|&|[g]|&&&
 &&&|[b]|&|[g]|&|[b]|&|[b]|&|[g]|&|[b]|&&&
 &&&|[r]|&|[r]|&|[g]|&|[g]|&|[r]|&|[r]|&&&
 &&&|[b]|&|[r]|&|[v]|&|[b]|&|[g]|&|[v]|&&\\
 &&&&&|[b]|&|[b]|&&&&&
 &&&&&|[b]|&|[g]|&&&&&
 &&&&&|[g]|&|[r]|&&&&&
 &&&&&|[b]|&|[g]|&&&&&
 &&&&&|[g]|&|[b]|&&&&&
 &&&&&|[g]|&|[r]|&&&&&
 &&&&&|[g]|&|[r]|&&&&&
 &&&&&|[g]|&|[r]|&&&&&
 &&&&&|[b]|&|[v]|&&&&&
 &&&&&|[b]|&|[b]|&&&&&
 &&&&&|[b]|&|[b]|&&&&&
 &&&&&|[g]|&|[r]|&&&&\\
 &&&&&|[b]|&|[b]|&&&&&
 &&&&&|[b]|&|[g]|&&&&&
 &&&&&|[g]|&|[r]|&&&&&
 &&&&&|[b]|&|[g]|&&&&&
 &&&&&|[g]|&|[b]|&&&&&
 &&&&&|[g]|&|[r]|&&&&&
 &&&&&|[r]|&|[g]|&&&&&
 &&&&&|[r]|&|[b]|&&&&&
 &&&&&|[r]|&|[g]|&&&&&
 &&&&&|[g]|&|[g]|&&&&&
 &&&&&|[g]|&|[g]|&&&&&
 &&&&&|[r]|&|[g]|&&&&\\&&&&&&&&&&&\phantom{}&&&&&&&&&&&\phantom{}&&&&&&&&&&&\phantom{}&&&&&&&&&&&\phantom{}&&&&&&&&&&&\phantom{}&&&&&&&&&&&\phantom{}&&&&&&&&&&&\phantom{}&&&&&&&&&&&\phantom{}&&&&&&&&&&&\phantom{}&&&&&&&&&&&\phantom{}&&&&&&&&&&&\phantom{}&&&&&&&&&&&\phantom{}\\
&|[l10]|&|[l10]|&|[l10]|&|[l10]|&|[l10]|&|[l10]|&|[l10]|&|[l10]|&|[l10]|&&&
 |[l10]|&|[l10]|&|[l10]|&|[l10]|&|[l10]|&|[l10]|&|[l10]|&|[l10]|&|[l10]|&&&
 |[l8]|&|[l8]|&|[l8]|&|[l8]|&|[l8]|&|[l8]|&|[l8]|&|[l8]|&|[l8]|&&&
 |[l8]|&|[l8]|&|[l8]|&|[l8]|&|[l8]|&|[l8]|&|[l8]|&|[l8]|&|[l8]|&&&
 |[l8]|&|[l8]|&|[l8]|&|[l8]|&|[l8]|&|[l8]|&|[l8]|&|[l8]|&|[l8]|&&&
 |[l8]|&|[l8]|&|[l8]|&|[l8]|&|[l8]|&|[l8]|&|[l8]|&|[l8]|&|[l8]|&&&
 |[l6]|&|[l6]|&|[l6]|&|[l6]|&|[l6]|&|[l6]|&|[l6]|&|[l6]|&|[l6]|&&&
 |[l6]|&|[l6]|&|[l6]|&|[l6]|&|[l6]|&|[l6]|&|[l6]|&|[l6]|&|[l6]|&&&
 |[l6]|&|[l6]|&|[l6]|&|[l6]|&|[l6]|&|[l6]|&|[l6]|&|[l6]|&|[l6]|&&&
 |[l6]|&|[l6]|&|[l6]|&|[l6]|&|[l6]|&|[l6]|&|[l6]|&|[l6]|&|[l6]|&&&
 |[l6]|&|[l6]|&|[l6]|&|[l6]|&|[l6]|&|[l6]|&|[l6]|&|[l6]|&|[l6]|&&&
 |[l4]|&|[l4]|&|[l4]|&|[l4]|&|[l4]|&|[l4]|&|[l4]|&|[l4]|&|[l4]|&&\\
&|[n]|&|[l9]|&|[l9]|&|[n]|&|[l9]|&|[l9]|&|[n]|&|[l9]|&|[l9]|&&&
 |[n]|&|[l9]|&|[l9]|&|[n]|&|[l9]|&|[l9]|&|[n]|&|[l9]|&|[l9]|&&&
 |[n]|&|[l7]|&|[l7]|&|[n]|&|[l7]|&|[l7]|&|[n]|&|[l7]|&|[l7]|&&&
 |[n]|&|[l7]|&|[l7]|&|[n]|&|[l7]|&|[l7]|&|[n]|&|[l7]|&|[l7]|&&&
 |[n]|&|[l7]|&|[l7]|&|[n]|&|[l7]|&|[l7]|&|[n]|&|[l7]|&|[l7]|&&&
 |[n]|&|[l7]|&|[l7]|&|[n]|&|[l7]|&|[l7]|&|[n]|&|[l7]|&|[l7]|&&&
 |[n]|&|[l5]|&|[l5]|&|[n]|&|[l5]|&|[l5]|&|[n]|&|[l5]|&|[l5]|&&&
 |[n]|&|[l5]|&|[l5]|&|[n]|&|[l5]|&|[l5]|&|[n]|&|[l5]|&|[l5]|&&&
 |[n]|&|[l5]|&|[l5]|&|[n]|&|[l5]|&|[l5]|&|[n]|&|[l5]|&|[l5]|&&&
 |[n]|&|[l5]|&|[l5]|&|[n]|&|[l5]|&|[l5]|&|[n]|&|[l5]|&|[l5]|&&&
 |[n]|&|[l5]|&|[l5]|&|[n]|&|[l5]|&|[l5]|&|[n]|&|[l5]|&|[l5]|&&&
 |[n]|&|[l3]|&|[l3]|&|[n]|&|[l3]|&|[l3]|&|[n]|&|[l3]|&|[l3]|&&\\
&|[n]|&|[n]|&|[l8]|&|[n]|&|[n]|&|[l8]|&|[n]|&|[n]|&|[l8]|&&&
 |[n]|&|[n]|&|[l8]|&|[n]|&|[n]|&|[l8]|&|[n]|&|[n]|&|[l8]|&&&
 |[n]|&|[n]|&|[l6]|&|[n]|&|[n]|&|[l6]|&|[n]|&|[n]|&|[l6]|&&&
 |[n]|&|[n]|&|[l6]|&|[n]|&|[n]|&|[l6]|&|[n]|&|[n]|&|[l6]|&&&
 |[n]|&|[n]|&|[l6]|&|[n]|&|[n]|&|[l6]|&|[n]|&|[n]|&|[l6]|&&&
 |[n]|&|[n]|&|[l6]|&|[n]|&|[n]|&|[l6]|&|[n]|&|[n]|&|[l6]|&&&
 |[n]|&|[n]|&|[l4]|&|[n]|&|[n]|&|[l4]|&|[n]|&|[n]|&|[l4]|&&&
 |[n]|&|[n]|&|[l4]|&|[n]|&|[n]|&|[l4]|&|[n]|&|[n]|&|[l4]|&&&
 |[n]|&|[n]|&|[l4]|&|[n]|&|[n]|&|[l4]|&|[n]|&|[n]|&|[l4]|&&&
 |[n]|&|[n]|&|[l4]|&|[n]|&|[n]|&|[l4]|&|[n]|&|[n]|&|[l4]|&&&
 |[n]|&|[n]|&|[l4]|&|[n]|&|[n]|&|[l4]|&|[n]|&|[n]|&|[l4]|&&&
 |[n]|&|[n]|&|[l2]|&|[n]|&|[n]|&|[l2]|&|[n]|&|[n]|&|[l2]|&&\\
&|[n]|&|[n]|&|[n]|&|[l9]|&|[l9]|&|[l9]|&|[l9]|&|[l9]|&|[l9]|&&&
 |[n]|&|[n]|&|[n]|&|[l9]|&|[l9]|&|[l9]|&|[l9]|&|[l9]|&|[l9]|&&&
 |[n]|&|[n]|&|[n]|&|[l7]|&|[l7]|&|[l7]|&|[l7]|&|[l7]|&|[l7]|&&&
 |[n]|&|[n]|&|[n]|&|[l7]|&|[l7]|&|[l7]|&|[l7]|&|[l7]|&|[l7]|&&&
 |[n]|&|[n]|&|[n]|&|[l7]|&|[l7]|&|[l7]|&|[l7]|&|[l7]|&|[l7]|&&&
 |[n]|&|[n]|&|[n]|&|[l7]|&|[l7]|&|[l7]|&|[l7]|&|[l7]|&|[l7]|&&&
 |[n]|&|[n]|&|[n]|&|[l5]|&|[l5]|&|[l5]|&|[l5]|&|[l5]|&|[l5]|&&&
 |[n]|&|[n]|&|[n]|&|[l5]|&|[l5]|&|[l5]|&|[l5]|&|[l5]|&|[l5]|&&&
 |[n]|&|[n]|&|[n]|&|[l5]|&|[l5]|&|[l5]|&|[l5]|&|[l5]|&|[l5]|&&&
 |[n]|&|[n]|&|[n]|&|[l5]|&|[l5]|&|[l5]|&|[l5]|&|[l5]|&|[l5]|&&&
 |[n]|&|[n]|&|[n]|&|[l5]|&|[l5]|&|[l5]|&|[l5]|&|[l5]|&|[l5]|&&&
 |[n]|&|[n]|&|[n]|&|[l3]|&|[l3]|&|[l3]|&|[l3]|&|[l3]|&|[l3]|&&\\
&|[n]|&|[n]|&|[n]|&|[n]|&|[l8]|&|[l8]|&|[n]|&|[l8]|&|[l8]|&&&
 |[n]|&|[n]|&|[n]|&|[n]|&|[l8]|&|[l8]|&|[n]|&|[l8]|&|[l8]|&&&
 |[n]|&|[n]|&|[n]|&|[n]|&|[l6]|&|[l6]|&|[n]|&|[l6]|&|[l6]|&&&
 |[n]|&|[n]|&|[n]|&|[n]|&|[l6]|&|[l6]|&|[n]|&|[l6]|&|[l6]|&&&
 |[n]|&|[n]|&|[n]|&|[n]|&|[l6]|&|[l6]|&|[n]|&|[l6]|&|[l6]|&&&
 |[n]|&|[n]|&|[n]|&|[n]|&|[l6]|&|[l6]|&|[n]|&|[l6]|&|[l6]|&&&
 |[n]|&|[n]|&|[n]|&|[n]|&|[l4]|&|[l4]|&|[n]|&|[l4]|&|[l4]|&&&
 |[n]|&|[n]|&|[n]|&|[n]|&|[l4]|&|[l4]|&|[n]|&|[l4]|&|[l4]|&&&
 |[n]|&|[n]|&|[n]|&|[n]|&|[l4]|&|[l4]|&|[n]|&|[l4]|&|[l4]|&&&
 |[n]|&|[n]|&|[n]|&|[n]|&|[l4]|&|[l4]|&|[n]|&|[l4]|&|[l4]|&&&
 |[n]|&|[n]|&|[n]|&|[n]|&|[l4]|&|[l4]|&|[n]|&|[l4]|&|[l4]|&&&
 |[n]|&|[n]|&|[n]|&|[n]|&|[l2]|&|[l2]|&|[n]|&|[l2]|&|[l2]|&&\\
&|[n]|&|[n]|&|[n]|&|[n]|&|[n]|&|[l7]|&|[n]|&|[n]|&|[l7]|&&&
 |[n]|&|[n]|&|[n]|&|[n]|&|[n]|&|[l7]|&|[n]|&|[n]|&|[l7]|&&&
 |[n]|&|[n]|&|[n]|&|[n]|&|[n]|&|[l5]|&|[n]|&|[n]|&|[l5]|&&&
 |[n]|&|[n]|&|[n]|&|[n]|&|[n]|&|[l5]|&|[n]|&|[n]|&|[l5]|&&&
 |[n]|&|[n]|&|[n]|&|[n]|&|[n]|&|[l5]|&|[n]|&|[n]|&|[l5]|&&&
 |[n]|&|[n]|&|[n]|&|[n]|&|[n]|&|[l5]|&|[n]|&|[n]|&|[l5]|&&&
 |[n]|&|[n]|&|[n]|&|[n]|&|[n]|&|[l3]|&|[n]|&|[n]|&|[l3]|&&&
 |[n]|&|[n]|&|[n]|&|[n]|&|[n]|&|[l3]|&|[n]|&|[n]|&|[l3]|&&&
 |[n]|&|[n]|&|[n]|&|[n]|&|[n]|&|[l3]|&|[n]|&|[n]|&|[l3]|&&&
 |[n]|&|[n]|&|[n]|&|[n]|&|[n]|&|[l3]|&|[n]|&|[n]|&|[l3]|&&&
 |[n]|&|[n]|&|[n]|&|[n]|&|[n]|&|[l3]|&|[n]|&|[n]|&|[l3]|&&&
 |[n]|&|[n]|&|[n]|&|[n]|&|[n]|&|[l1]|&|[n]|&|[n]|&|[l1]|&&\\
&|[n]|&|[n]|&|[n]|&|[n]|&|[n]|&|[n]|&|[l8]|&|[l8]|&|[l8]|&&&
 |[n]|&|[n]|&|[n]|&|[n]|&|[n]|&|[n]|&|[l8]|&|[l8]|&|[l8]|&&&
 |[n]|&|[n]|&|[n]|&|[n]|&|[n]|&|[n]|&|[l6]|&|[l6]|&|[l6]|&&&
 |[n]|&|[n]|&|[n]|&|[n]|&|[n]|&|[n]|&|[l6]|&|[l6]|&|[l6]|&&&
 |[n]|&|[n]|&|[n]|&|[n]|&|[n]|&|[n]|&|[l6]|&|[l6]|&|[l6]|&&&
 |[n]|&|[n]|&|[n]|&|[n]|&|[n]|&|[n]|&|[l6]|&|[l6]|&|[l6]|&&&
 |[n]|&|[n]|&|[n]|&|[n]|&|[n]|&|[n]|&|[l4]|&|[l4]|&|[l4]|&&&
 |[n]|&|[n]|&|[n]|&|[n]|&|[n]|&|[n]|&|[l4]|&|[l4]|&|[l4]|&&&
 |[n]|&|[n]|&|[n]|&|[n]|&|[n]|&|[n]|&|[l4]|&|[l4]|&|[l4]|&&&
 |[n]|&|[n]|&|[n]|&|[n]|&|[n]|&|[n]|&|[l4]|&|[l4]|&|[l4]|&&&
 |[n]|&|[n]|&|[n]|&|[n]|&|[n]|&|[n]|&|[l4]|&|[l4]|&|[l4]|&&&
 |[n]|&|[n]|&|[n]|&|[n]|&|[n]|&|[n]|&|[l2]|&|[l2]|&|[l2]|&&\\
&|[n]|&|[n]|&|[n]|&|[n]|&|[n]|&|[n]|&|[n]|&|[l7]|&|[l7]|&&&
 |[n]|&|[n]|&|[n]|&|[n]|&|[n]|&|[n]|&|[n]|&|[l7]|&|[l7]|&&&
 |[n]|&|[n]|&|[n]|&|[n]|&|[n]|&|[n]|&|[n]|&|[l5]|&|[l5]|&&&
 |[n]|&|[n]|&|[n]|&|[n]|&|[n]|&|[n]|&|[n]|&|[l5]|&|[l5]|&&&
 |[n]|&|[n]|&|[n]|&|[n]|&|[n]|&|[n]|&|[n]|&|[l5]|&|[l5]|&&&
 |[n]|&|[n]|&|[n]|&|[n]|&|[n]|&|[n]|&|[n]|&|[l5]|&|[l5]|&&&
 |[n]|&|[n]|&|[n]|&|[n]|&|[n]|&|[n]|&|[n]|&|[l3]|&|[l3]|&&&
 |[n]|&|[n]|&|[n]|&|[n]|&|[n]|&|[n]|&|[n]|&|[l3]|&|[l3]|&&&
 |[n]|&|[n]|&|[n]|&|[n]|&|[n]|&|[n]|&|[n]|&|[l3]|&|[l3]|&&&
 |[n]|&|[n]|&|[n]|&|[n]|&|[n]|&|[n]|&|[n]|&|[l3]|&|[l3]|&&&
 |[n]|&|[n]|&|[n]|&|[n]|&|[n]|&|[n]|&|[n]|&|[l3]|&|[l3]|&&&
 |[n]|&|[n]|&|[n]|&|[n]|&|[n]|&|[n]|&|[n]|&|[l1]|&|[l1]|&&\\
&|[n]|&|[n]|&|[n]|&|[n]|&|[n]|&|[n]|&|[n]|&|[n]|&|[l6]|&&&
 |[n]|&|[n]|&|[n]|&|[n]|&|[n]|&|[n]|&|[n]|&|[n]|&|[l6]|&&&
 |[n]|&|[n]|&|[n]|&|[n]|&|[n]|&|[n]|&|[n]|&|[n]|&|[l4]|&&&
 |[n]|&|[n]|&|[n]|&|[n]|&|[n]|&|[n]|&|[n]|&|[n]|&|[l4]|&&&
 |[n]|&|[n]|&|[n]|&|[n]|&|[n]|&|[n]|&|[n]|&|[n]|&|[l4]|&&&
 |[n]|&|[n]|&|[n]|&|[n]|&|[n]|&|[n]|&|[n]|&|[n]|&|[l4]|&&&
 |[n]|&|[n]|&|[n]|&|[n]|&|[n]|&|[n]|&|[n]|&|[n]|&|[l2]|&&&
 |[n]|&|[n]|&|[n]|&|[n]|&|[n]|&|[n]|&|[n]|&|[n]|&|[l2]|&&&
 |[n]|&|[n]|&|[n]|&|[n]|&|[n]|&|[n]|&|[n]|&|[n]|&|[l2]|&&&
 |[n]|&|[n]|&|[n]|&|[n]|&|[n]|&|[n]|&|[n]|&|[n]|&|[l2]|&&&
 |[n]|&|[n]|&|[n]|&|[n]|&|[n]|&|[n]|&|[n]|&|[n]|&|[l2]|&&&
 |[n]|&|[n]|&|[n]|&|[n]|&|[n]|&|[n]|&|[n]|&|[n]|&|[l0]|&&\\
 &&&&&&&&&&&\phantom{}&&&&&&&&&&&\phantom{}&&&&&&&&&&&\phantom{}&&&&&&&&&&&\phantom{}&&&&&&&&&&&\phantom{}&&&&&&&&&&&\phantom{}&&&&&&&&&&&\phantom{}&&&&&&&&&&&\phantom{}&&&&&&&&&&&\phantom{}&&&&&&&&&&&\phantom{}&&&&&&&&&&&\phantom{}&&&&&&&&&&&\phantom{}\\
&|[l10]|&|[n]|&|[n]|&|[n]|&|[n]|&|[n]|&|[n]|&|[n]|&|[n]|&|[n]|&&
 |[l10]|&|[n]|&|[n]|&|[n]|&|[n]|&|[n]|&|[n]|&|[n]|&|[n]|&|[n]|&&
 |[l8]|&|[n]|&|[n]|&|[n]|&|[n]|&|[n]|&|[n]|&|[n]|&|[n]|&|[n]|&&
 |[l8]|&|[n]|&|[n]|&|[n]|&|[n]|&|[n]|&|[n]|&|[n]|&|[n]|&|[n]|&&
 |[l8]|&|[n]|&|[n]|&|[n]|&|[n]|&|[n]|&|[n]|&|[n]|&|[n]|&|[n]|&&
 |[l8]|&|[n]|&|[n]|&|[n]|&|[n]|&|[n]|&|[n]|&|[n]|&|[n]|&|[n]|&&
 |[l6]|&|[n]|&|[n]|&|[n]|&|[n]|&|[n]|&|[n]|&|[n]|&|[n]|&|[n]|&&
 |[l6]|&|[n]|&|[n]|&|[n]|&|[n]|&|[n]|&|[n]|&|[n]|&|[n]|&|[n]|&&
 |[l6]|&|[n]|&|[n]|&|[n]|&|[n]|&|[n]|&|[n]|&|[n]|&|[n]|&|[n]|&&
 |[l6]|&|[n]|&|[n]|&|[n]|&|[n]|&|[n]|&|[n]|&|[n]|&|[n]|&|[n]|&&
 |[l6]|&|[n]|&|[n]|&|[n]|&|[n]|&|[n]|&|[n]|&|[n]|&|[n]|&|[n]|&&
 |[l4]|&|[n]|&|[n]|&|[n]|&|[n]|&|[n]|&|[n]|&|[n]|&|[n]|&|[n]|&\\
&|[l9]|&|[n]|&|[n]|&|[n]|&|[n]|&|[n]|&|[n]|&|[n]|&|[n]|&|[n]|&&
 |[l9]|&|[n]|&|[n]|&|[n]|&|[n]|&|[n]|&|[n]|&|[n]|&|[n]|&|[n]|&&
 |[l7]|&|[n]|&|[n]|&|[n]|&|[n]|&|[n]|&|[n]|&|[n]|&|[n]|&|[n]|&&
 |[l7]|&|[n]|&|[n]|&|[n]|&|[n]|&|[n]|&|[n]|&|[n]|&|[n]|&|[n]|&&
 |[l7]|&|[n]|&|[n]|&|[n]|&|[n]|&|[n]|&|[n]|&|[n]|&|[n]|&|[n]|&&
 |[l7]|&|[n]|&|[n]|&|[n]|&|[n]|&|[n]|&|[n]|&|[n]|&|[n]|&|[n]|&&
 |[l5]|&|[n]|&|[n]|&|[n]|&|[n]|&|[n]|&|[n]|&|[n]|&|[n]|&|[n]|&&
 |[l5]|&|[n]|&|[n]|&|[n]|&|[n]|&|[n]|&|[n]|&|[n]|&|[n]|&|[n]|&&
 |[l5]|&|[n]|&|[n]|&|[n]|&|[n]|&|[n]|&|[n]|&|[n]|&|[n]|&|[n]|&&
 |[l5]|&|[n]|&|[n]|&|[n]|&|[n]|&|[n]|&|[n]|&|[n]|&|[n]|&|[n]|&&
 |[l5]|&|[n]|&|[n]|&|[n]|&|[n]|&|[n]|&|[n]|&|[n]|&|[n]|&|[n]|&&
 |[l3]|&|[n]|&|[n]|&|[n]|&|[n]|&|[n]|&|[n]|&|[n]|&|[n]|&|[n]|&\\
&|[l8]|&|[n]|&|[n]|&|[n]|&|[n]|&|[n]|&|[n]|&|[n]|&|[n]|&|[n]|&&
 |[l8]|&|[n]|&|[n]|&|[n]|&|[n]|&|[n]|&|[n]|&|[n]|&|[n]|&|[n]|&&
 |[l6]|&|[n]|&|[n]|&|[n]|&|[n]|&|[n]|&|[n]|&|[n]|&|[n]|&|[n]|&&
 |[l6]|&|[n]|&|[n]|&|[n]|&|[n]|&|[n]|&|[n]|&|[n]|&|[n]|&|[n]|&&
 |[l6]|&|[n]|&|[n]|&|[n]|&|[n]|&|[n]|&|[n]|&|[n]|&|[n]|&|[n]|&&
 |[l6]|&|[n]|&|[n]|&|[n]|&|[n]|&|[n]|&|[n]|&|[n]|&|[n]|&|[n]|&&
 |[l4]|&|[n]|&|[n]|&|[n]|&|[n]|&|[n]|&|[n]|&|[n]|&|[n]|&|[n]|&&
 |[l4]|&|[n]|&|[n]|&|[n]|&|[n]|&|[n]|&|[n]|&|[n]|&|[n]|&|[n]|&&
 |[l4]|&|[n]|&|[n]|&|[n]|&|[n]|&|[n]|&|[n]|&|[n]|&|[n]|&|[n]|&&
 |[l4]|&|[n]|&|[n]|&|[n]|&|[n]|&|[n]|&|[n]|&|[n]|&|[n]|&|[n]|&&
 |[l4]|&|[n]|&|[n]|&|[n]|&|[n]|&|[n]|&|[n]|&|[n]|&|[n]|&|[n]|&&
 |[l2]|&|[n]|&|[n]|&|[n]|&|[n]|&|[n]|&|[n]|&|[n]|&|[n]|&|[n]|&\\
&|[l9]|&|[l9]|&|[l9]|&|[l9]|&|[l9]|&|[l9]|&|[l9]|&|[n]|&|[n]|&|[n]|&&
 |[l9]|&|[l9]|&|[l9]|&|[l9]|&|[l9]|&|[l9]|&|[l9]|&|[n]|&|[n]|&|[n]|&&
 |[l7]|&|[l7]|&|[l7]|&|[n]|&|[l7]|&|[n]|&|[n]|&|[n]|&|[n]|&|[n]|&&
 |[l7]|&|[l7]|&|[l7]|&|[n]|&|[n]|&|[n]|&|[n]|&|[n]|&|[n]|&|[n]|&&
 |[l7]|&|[l7]|&|[l7]|&|[l7]|&|[l7]|&|[n]|&|[n]|&|[n]|&|[n]|&|[n]|&&
 |[l7]|&|[l7]|&|[l7]|&|[n]|&|[l7]|&|[l7]|&|[n]|&|[n]|&|[n]|&|[n]|&&
 |[l5]|&|[l5]|&|[n]|&|[n]|&|[l5]|&|[n]|&|[l5]|&|[n]|&|[n]|&|[n]|&&
 |[l5]|&|[l5]|&|[n]|&|[n]|&|[l5]|&|[n]|&|[l5]|&|[n]|&|[n]|&|[n]|&&
 |[l5]|&|[l5]|&|[n]|&|[n]|&|[l5]|&|[n]|&|[n]|&|[n]|&|[n]|&|[n]|&&
 |[l5]|&|[l5]|&|[l5]|&|[l5]|&|[l5]|&|[n]|&|[l5]|&|[n]|&|[n]|&|[n]|&&
 |[l5]|&|[l5]|&|[l5]|&|[n]|&|[l5]|&|[n]|&|[n]|&|[n]|&|[n]|&|[n]|&&
 |[l3]|&|[l3]|&|[n]|&|[n]|&|[l3]|&|[n]|&|[l3]|&|[n]|&|[n]|&|[n]|&\\
&|[l8]|&|[l8]|&|[l8]|&|[l8]|&|[l8]|&|[l8]|&|[l8]|&|[n]|&|[n]|&|[n]|&&
 |[l8]|&|[l8]|&|[l8]|&|[l8]|&|[l8]|&|[l8]|&|[l8]|&|[n]|&|[n]|&|[n]|&&
 |[l6]|&|[l6]|&|[l6]|&|[n]|&|[l6]|&|[n]|&|[n]|&|[n]|&|[n]|&|[n]|&&
 |[l6]|&|[l6]|&|[l6]|&|[n]|&|[n]|&|[n]|&|[n]|&|[n]|&|[n]|&|[n]|&&
 |[l6]|&|[l6]|&|[l6]|&|[l6]|&|[l6]|&|[n]|&|[n]|&|[n]|&|[n]|&|[n]|&&
 |[l6]|&|[l6]|&|[l6]|&|[n]|&|[l6]|&|[l6]|&|[n]|&|[n]|&|[n]|&|[n]|&&
 |[l4]|&|[l4]|&|[n]|&|[n]|&|[l4]|&|[n]|&|[l4]|&|[n]|&|[n]|&|[n]|&&
 |[l4]|&|[l4]|&|[n]|&|[n]|&|[l4]|&|[n]|&|[l4]|&|[n]|&|[n]|&|[n]|&&
 |[l4]|&|[l4]|&|[n]|&|[n]|&|[l4]|&|[n]|&|[n]|&|[n]|&|[n]|&|[n]|&&
 |[l4]|&|[l4]|&|[l4]|&|[l4]|&|[l4]|&|[n]|&|[l4]|&|[n]|&|[n]|&|[n]|&&
 |[l4]|&|[l4]|&|[l4]|&|[n]|&|[l4]|&|[n]|&|[n]|&|[n]|&|[n]|&|[n]|&&
 |[l2]|&|[l2]|&|[n]|&|[n]|&|[l2]|&|[n]|&|[l2]|&|[n]|&|[n]|&|[n]|&\\
&|[l7]|&|[l7]|&|[l7]|&|[l7]|&|[l7]|&|[l7]|&|[l7]|&|[n]|&|[n]|&|[n]|&&
 |[l7]|&|[l7]|&|[l7]|&|[l7]|&|[l7]|&|[l7]|&|[l7]|&|[n]|&|[n]|&|[n]|&&
 |[l5]|&|[l5]|&|[l5]|&|[n]|&|[l5]|&|[n]|&|[n]|&|[n]|&|[n]|&|[n]|&&
 |[l5]|&|[l5]|&|[l5]|&|[n]|&|[n]|&|[n]|&|[n]|&|[n]|&|[n]|&|[n]|&&
 |[l5]|&|[l5]|&|[l5]|&|[l5]|&|[l5]|&|[n]|&|[n]|&|[n]|&|[n]|&|[n]|&&
 |[l5]|&|[l5]|&|[l5]|&|[n]|&|[l5]|&|[l5]|&|[n]|&|[n]|&|[n]|&|[n]|&&
 |[l3]|&|[l3]|&|[n]|&|[n]|&|[l3]|&|[n]|&|[l3]|&|[n]|&|[n]|&|[n]|&&
 |[l3]|&|[l3]|&|[n]|&|[n]|&|[l3]|&|[n]|&|[l3]|&|[n]|&|[n]|&|[n]|&&
 |[l3]|&|[l3]|&|[n]|&|[n]|&|[l3]|&|[n]|&|[n]|&|[n]|&|[n]|&|[n]|&&
 |[l3]|&|[l3]|&|[l3]|&|[l3]|&|[l3]|&|[n]|&|[l3]|&|[n]|&|[n]|&|[n]|&&
 |[l3]|&|[l3]|&|[l3]|&|[n]|&|[l3]|&|[n]|&|[n]|&|[n]|&|[n]|&|[n]|&&
 |[l1]|&|[l1]|&|[n]|&|[n]|&|[l1]|&|[n]|&|[l1]|&|[n]|&|[n]|&|[n]|&\\
&|[l8]|&|[l8]|&|[l8]|&|[l8]|&|[l8]|&|[l8]|&|[l8]|&|[n]|&|[n]|&|[l8]|&&
 |[l8]|&|[l8]|&|[l8]|&|[l8]|&|[l8]|&|[l8]|&|[l8]|&|[l8]|&|[l8]|&|[l8]|&&
 |[l6]|&|[l6]|&|[l6]|&|[n]|&|[l6]|&|[n]|&|[n]|&|[l6]|&|[l6]|&|[l6]|&&
 |[l6]|&|[l6]|&|[l6]|&|[n]|&|[n]|&|[n]|&|[n]|&|[l6]|&|[l6]|&|[l6]|&&
 |[l6]|&|[l6]|&|[l6]|&|[l6]|&|[l6]|&|[n]|&|[n]|&|[l6]|&|[l6]|&|[l6]|&&
 |[l6]|&|[l6]|&|[l6]|&|[n]|&|[l6]|&|[l6]|&|[n]|&|[l6]|&|[l6]|&|[l6]|&&
 |[l4]|&|[l4]|&|[n]|&|[n]|&|[l4]|&|[n]|&|[l4]|&|[l4]|&|[n]|&|[l4]|&&
 |[l4]|&|[l4]|&|[n]|&|[n]|&|[l4]|&|[n]|&|[l4]|&|[l4]|&|[l4]|&|[l4]|&&
 |[l4]|&|[l4]|&|[n]|&|[n]|&|[l4]|&|[n]|&|[n]|&|[l4]|&|[l4]|&|[l4]|&&
 |[l4]|&|[l4]|&|[l4]|&|[l4]|&|[l4]|&|[n]|&|[l4]|&|[l4]|&|[l4]|&|[l4]|&&
 |[l4]|&|[l4]|&|[l4]|&|[n]|&|[l4]|&|[n]|&|[n]|&|[l4]|&|[l4]|&|[l4]|&&
 |[l2]|&|[l2]|&|[n]|&|[n]|&|[l2]|&|[n]|&|[l2]|&|[n]|&|[l2]|&|[l2]|&\\
&|[l7]|&|[l7]|&|[l7]|&|[l7]|&|[l7]|&|[l7]|&|[l7]|&|[n]|&|[n]|&|[l7]|&&
 |[l7]|&|[l7]|&|[l7]|&|[l7]|&|[l7]|&|[l7]|&|[l7]|&|[l7]|&|[l7]|&|[l7]|&&
 |[l5]|&|[l5]|&|[l5]|&|[n]|&|[l5]|&|[n]|&|[n]|&|[l5]|&|[l5]|&|[l5]|&&
 |[l5]|&|[l5]|&|[l5]|&|[n]|&|[n]|&|[n]|&|[n]|&|[l5]|&|[l5]|&|[l5]|&&
 |[l5]|&|[l5]|&|[l5]|&|[l5]|&|[l5]|&|[n]|&|[n]|&|[l5]|&|[l5]|&|[l5]|&&
 |[l5]|&|[l5]|&|[l5]|&|[n]|&|[l5]|&|[l5]|&|[n]|&|[l5]|&|[l5]|&|[l5]|&&
 |[l3]|&|[l3]|&|[n]|&|[n]|&|[l3]|&|[n]|&|[l3]|&|[l3]|&|[n]|&|[l3]|&&
 |[l3]|&|[l3]|&|[n]|&|[n]|&|[l3]|&|[n]|&|[l3]|&|[l3]|&|[l3]|&|[l3]|&&
 |[l3]|&|[l3]|&|[n]|&|[n]|&|[l3]|&|[n]|&|[n]|&|[l3]|&|[l3]|&|[l3]|&&
 |[l3]|&|[l3]|&|[l3]|&|[l3]|&|[l3]|&|[n]|&|[l3]|&|[l3]|&|[l3]|&|[l3]|&&
 |[l3]|&|[l3]|&|[l3]|&|[n]|&|[l3]|&|[n]|&|[n]|&|[l3]|&|[l3]|&|[l3]|&&
 |[l1]|&|[l1]|&|[n]|&|[n]|&|[l1]|&|[n]|&|[l1]|&|[n]|&|[l1]|&|[l1]|&\\
&|[l6]|&|[l6]|&|[l6]|&|[l6]|&|[l6]|&|[l6]|&|[l6]|&|[n]|&|[n]|&|[l6]|&&
 |[l6]|&|[l6]|&|[l6]|&|[l6]|&|[l6]|&|[l6]|&|[l6]|&|[l6]|&|[l6]|&|[l6]|&&
 |[l4]|&|[l4]|&|[l4]|&|[n]|&|[l4]|&|[n]|&|[n]|&|[l4]|&|[l4]|&|[l4]|&&
 |[l4]|&|[l4]|&|[l4]|&|[n]|&|[n]|&|[n]|&|[n]|&|[l4]|&|[l4]|&|[l4]|&&
 |[l4]|&|[l4]|&|[l4]|&|[l4]|&|[l4]|&|[n]|&|[n]|&|[l4]|&|[l4]|&|[l4]|&&
 |[l4]|&|[l4]|&|[l4]|&|[n]|&|[l4]|&|[l4]|&|[n]|&|[l4]|&|[l4]|&|[l4]|&&
 |[l2]|&|[l2]|&|[n]|&|[n]|&|[l2]|&|[n]|&|[l2]|&|[l2]|&|[n]|&|[l2]|&&
 |[l2]|&|[l2]|&|[n]|&|[n]|&|[l2]|&|[n]|&|[l2]|&|[l2]|&|[l2]|&|[l2]|&&
 |[l2]|&|[l2]|&|[n]|&|[n]|&|[l2]|&|[n]|&|[n]|&|[l2]|&|[l2]|&|[l2]|&&
 |[l2]|&|[l2]|&|[l2]|&|[l2]|&|[l2]|&|[n]|&|[l2]|&|[l2]|&|[l2]|&|[l2]|&&
 |[l2]|&|[l2]|&|[l2]|&|[n]|&|[l2]|&|[n]|&|[n]|&|[l2]|&|[l2]|&|[l2]|&&
 |[l0]|&|[l0]|&|[n]|&|[n]|&|[l0]|&|[n]|&|[l0]|&|[n]|&|[l0]|&|[l0]|&\\
};
\end{tikzpicture}
\caption{
The 12 CAT schemes $T$ (top); the values of the inhibition parameters $\lambda^t_{rc}$ for the target skill nodes (centre); the value of the inhibition parameters $\lambda^t_{S_i}$ for the supplementary skill nodes (bottom). The inhibition parameters for both the target and supplementary skill are depicted as a matrix of nine rows representing the answers, and as many columns as the number of modelled skills. 
The strength of the skill-answer relation has eleven levels, from $0.1$ to $0.6$ with a step of $0.05$. Darker shades of grey mean lower skill-answer inhibition probabilities and white squares denote non-relevant skills.}
\label{fig:its}
\end{figure*}
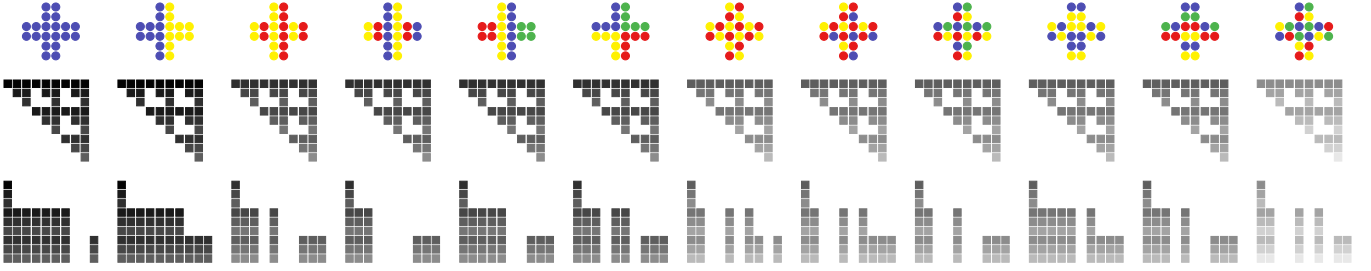

As a baseline summary measure used for checking the consistency of this score with the expert-based assessment in \cite{piatti_2022}, we define the \textit{CAT score} as the average of the individual scores assigned to the 12 CAT schemes solved by each student according to Table~\ref{tab:catscore}.  

\subsection{Parameters' elicitation}
Having defined the structure of the model for the CAT activity, it is necessary to set the values of the prior probabilities $\pi_{*}$, and the 12 inhibitors $\lambda^t_{*}$, $t = 1, \dots, 12$, for both the target and supplementary skills. 
As in \cite{softcom}, uniform prior probabilities, i.e., \begin{math} \pi_{rc} = 0.50 \end{math}, have been assigned to each skill. However, when conditioning given the constraints nodes $D_{rc, r'c'}=1$, their probabilities, before the observation of any answer node, become \begin{math} \pi_{11}=0.95,\, \pi_{12}=0.8,\, \pi_{13}=0.5,\, \pi_{21}=0.8,\newline \pi_{22}=0.5,\, \pi_{23}=0.2,\, \pi_{31}=0.5,\, \pi_{32}=0.2,\, \pi_{33}=0.05\end{math}.
Concerning the inhibition parameters, we consider two different models. 
Similarly to what was done in \cite{softcom}, in the first \emph{baseline} model, hereafter referred to as \textit{Model B}, all inhibitors are set to the same value. 
In contrast, a domain expert has elicited parameters in the second \emph{enhanced} model, hereafter referred to as \textit{Model E}.

\begin{figure}[!ht]
\centering
\begin{subfigure}
\centering
\scriptsize
\begin{tikzpicture}
\tikzset{b/.style={draw,circle,scale=1,blue!40!gray,fill}}
\tikzset{g/.style={draw,circle,scale=1,yellow,fill}}
\tikzset{r/.style={draw,circle,scale=1,red!80!gray,fill}}
\tikzset{v/.style={draw,circle,scale=1,green!40!gray,fill}}
\tikzset{l10/.style={draw,rectangle,scale=1,black!99,fill}}   
\tikzset{l9/.style={draw,rectangle,scale=1,black!90,fill}}    
\tikzset{l8/.style={draw,rectangle,scale=1,black!81,fill}}    
\tikzset{l7/.style={draw,rectangle,scale=1,black!72,fill}}    
\tikzset{l6/.style={draw,rectangle,scale=1,black!63,fill}}    
\tikzset{l5/.style={draw,rectangle,scale=1,black!54,fill}}    
\tikzset{l4/.style={draw,rectangle,scale=1,black!45,fill}}    
\tikzset{n/.style={draw,rectangle,scale=0.62,black!0,fill}}
\matrix[matrix of nodes,row sep=0mm,column sep=-.5mm]
{
&$X_{11}$&$X_{12}$&$X_{13}$&$X_{21}$&$X_{22}$&$X_{23}$&$X_{31}$&$X_{32}$&$X_{33}$\\
$Y_1$&|[l8]|&|[l8]|&|[l8]|&|[l8]|&|[l8]|&|[l8]|&|[l8]|&|[l8]|&|[l8]|\\
$Y_2$&|[n]|&|[l7]|&|[l7]|&|[n]|&|[l7]|&|[l7]|&|[n]|&|[l7]|&|[l7]|\\
$Y_3$&|[n]|&|[n]|&|[l6]|&|[n]|&|[n]|&|[l6]|&|[n]|&|[n]|&|[l6]|\\
$Y_4$&|[n]|&|[n]|&|[n]|&|[l7]|&|[l7]|&|[l7]|&|[l7]|&|[l7]|&|[l7]|\\
$Y_5$&|[n]|&|[n]|&|[n]|&|[n]|&|[l6]|&|[l6]|&|[n]|&|[l6]|&|[l6]|\\
$Y_6$&|[n]|&|[n]|&|[n]|&|[n]|&|[n]|&|[l5]|&|[n]|&|[n]|&|[l5]|\\
$Y_7$&|[n]|&|[n]|&|[n]|&|[n]|&|[n]|&|[n]|&|[l6]|&|[l6]|&|[l6]|\\
$Y_8$&|[n]|&|[n]|&|[n]|&|[n]|&|[n]|&|[n]|&|[n]|&|[l5]|&|[l5]|\\
$Y_9$&|[n]|&|[n]|&|[n]|&|[n]|&|[n]|&|[n]|&|[n]|&|[n]|&|[l4]|\\
&|[n]|&|[n]|&|[n]|&|[n]|&|[n]|&|[n]|&|[n]|&|[n]|&|[n]|\\
};          
\end{tikzpicture}
\footnotesize
\\ (i) Inhibition parameters for the target skills.
\end{subfigure}
~
\begin{subfigure}
\centering
\scriptsize
\begin{tikzpicture}
\tikzset{b/.style={draw,circle,scale=1,blue!40!gray,fill}}
\tikzset{g/.style={draw,circle,scale=1,yellow,fill}}
\tikzset{r/.style={draw,circle,scale=1,red!80!gray,fill}}
\tikzset{v/.style={draw,circle,scale=1,green!40!gray,fill}}
\tikzset{l10/.style={draw,rectangle,scale=1,black!99,fill}}   
\tikzset{l9/.style={draw,rectangle,scale=1,black!90,fill}}    
\tikzset{l8/.style={draw,rectangle,scale=1,black!81,fill}}    
\tikzset{l7/.style={draw,rectangle,scale=1,black!72,fill}}    
\tikzset{l6/.style={draw,rectangle,scale=1,black!63,fill}}    
\tikzset{l5/.style={draw,rectangle,scale=1,black!54,fill}}    
\tikzset{l4/.style={draw,rectangle,scale=1,black!45,fill}}    
\tikzset{n/.style={draw,rectangle,scale=0.62,black!0,fill}}
\matrix[matrix of nodes,row sep=0mm,column sep=-.2mm]
{
&$S_1$&$S_2$&$S_3$&$S_4$&$S_5$&$S_6$&$S_7$&$S_8$&$S_9$&$S_{10}$\\
$Y_1$&|[l8]|&|[n]|&|[n]|&|[n]|&|[n]|&|[n]|&|[n]|&|[n]|&|[n]|&|[n]|\\
$Y_2$&|[l7]|&|[n]|&|[n]|&|[n]|&|[n]|&|[n]|&|[n]|&|[n]|&|[n]|&|[n]|\\
$Y_3$&|[l6]|&|[n]|&|[n]|&|[n]|&|[n]|&|[n]|&|[n]|&|[n]|&|[n]|&|[n]|\\
$Y_4$&|[l7]|&|[l7]|&|[l7]|&|[n]|&|[l7]|&|[n]|&|[n]|&|[n]|&|[n]|&|[n]|\\
$Y_5$&|[l6]|&|[l6]|&|[l6]|&|[n]|&|[l6]|&|[n]|&|[n]|&|[n]|&|[n]|&|[n]|\\
$Y_6$&|[l5]|&|[l5]|&|[l5]|&|[n]|&|[l5]|&|[n]|&|[n]|&|[n]|&|[n]|&|[n]|\\
$Y_7$&|[l6]|&|[l6]|&|[l6]|&|[n]|&|[l6]|&|[n]|&|[n]|&|[l6]|&|[l6]|&|[l6]|\\
$Y_8$&|[l5]|&|[l5]|&|[l5]|&|[n]|&|[l5]|&|[n]|&|[n]|&|[l5]|&|[l5]|&|[l5]|\\
$Y_9$&|[l4]|&|[l4]|&|[l4]|&|[n]|&|[l4]|&|[n]|&|[n]|&|[l4]|&|[l4]|&|[l4]|\\
};          
\end{tikzpicture}
\footnotesize
\\ (ii) Inhibition parameters of the supplementary skills.
\end{subfigure}
\normalsize
\caption{
Inhibition parameters $\lambda^{T3}_{*}$ used in the ECS model for schema T3 (Zoom on schema T3 of Fig.~\ref{fig:its}). The parameters are divided into two sets: the target skills (top) and the supplementary ones (bottom). 
The supplementary skills $S_4$ (paint monochromatic squares), $S_6$ (paint monochromatic ls), and $S_7$ (paint monochromatic zigzags) are represented as empty columns because they cannot be used to solve task T3.}
\label{fig:sampleits}
\end{figure}

Model B may look trivial and unrealistic, but it allows one to understand better the effect of the constraints resulting from ordering the skills and supplementary skills on the model inferences. 
The constant value of $\lambda$ was chosen equal to $0.2$, except for the leak node, associated with all answer nodes and modelling a guess probability of $0.1$, resulting in  $\lambda_{\mathrm{leak}} = 0.9$.

Model E is an enhanced version of the baseline model designed to address the increasing difficulties of the 12 tasks and the challenges students may encounter applying their skills to different schemes. 
The expert elicitation process involved grouping the 12 schemes into eight categories of increasing difficulties based on their characteristics: (i) T1, (ii) T2, (iii) T3, T4, (iv) T5, T6, (v) T7, T8, T9, (vi) T10, (vii) T11, (viii) T12. 
The expert assumed all tasks could be solved with 0D, 1D, and 2D algorithms. Moreover, given a schema $t$ and a manifest variable $Y^t_{rc}$, the same inhibition probability was assumed for all relevant skills, meaning that all have the same probability of successfully being applied in solving scheme $t$ with level $rc$. 
In the proposed method, the inhibitor parameter $\lambda_{rc}$ is used to model the probability of failing a task of a particular difficulty level $rc$, assuming the student has the necessary skills to solve the task. 
When a task is more complex or less help is available to the student, the value of $rc$ increases, which means that the inhibitor parameter also increases. 
This is because when the student possesses the necessary skills to solve a difficult task, the probability of failing is higher than for a simpler task.
Similarly, the inhibitor parameter $\lambda^t_{rc}$ is assigned to a particular scheme $t$ and is used to model the difficulty of implementing a solution of level $rc$ for that scheme. 
A high value of $\lambda^t_{rc}$ means that it is difficult to implement a solution of level $rc$ for that scheme. In other words, the inhibitor parameter $\lambda^t_{rc}$ provides a measure of the difficulty of implementing a particular solution for a given scheme at a given level of complexity.

While the students are generally expected to use 2D algorithms to solve the tasks optimally, there may be cases where a simpler 2D solution may be optimal.
Nonetheless, in the current implementation, the first two tasks are designed to serve as starting points for students, introducing them to the activity. They are expected to be solved using simpler 1D algorithms.
However, this particular case was not included in our model. This could have been described by setting high inhibitor values to indicate that certain 2D solutions are more difficult to implement than others, making them less likely to be chosen by students.

Our succinct elicitation setup allows summarising both the BN topology and its parameter values graphically (see the monochromatic rows at the bottom of Fig.~\ref{fig:its}) and explained more in detail in Fig.~\ref{fig:sampleits} for CAT schema T3. 

The underlying BN has been implemented within the CREMA Java library \cite{huber2020a}, which supports the specifications of noisy gates and inference based on these parametric CPTs. The network size allowed for exact inferences using the Variable Elimination algorithm \cite{chavira2007compiling}.

\begin{figure*}[!ht]
 \centering
 \includegraphics[width=.7\textwidth]{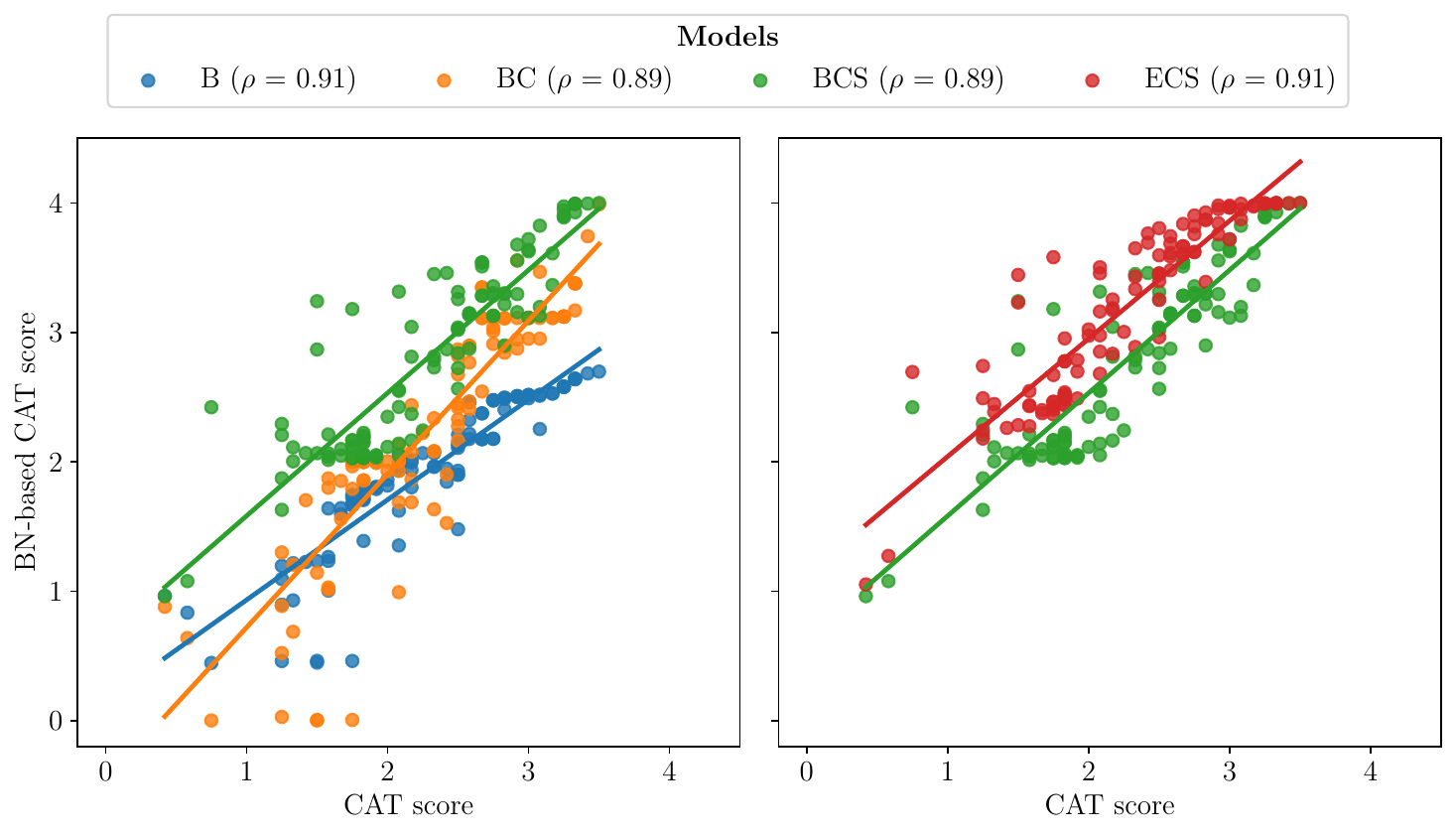}
  \caption{Scatterplot of the BN-based CAT scores versus the CAT score from \cite{piatti_2022} for the four models under comparison: \textit{B} — baseline proposed in \cite{softcom} –, \textit{BC} –baseline model with constraints –, BCS – baseline model with constraints and supplementary skills – and ECS – enhanced model with constraints and supplementary skills. 
  The legend gives the Pearson correlation coefficient $\rho$ between the two metrics.}
 \label{fig:scatter}
\end{figure*}

\section{Results} \label{sec:results} 
To evaluate the performance of our models, we processed the answers provided by all 109 pupils included in the study in \cite{piatti_2022}, obtaining for each model the posterior probabilities of the nine target skills and eventually for the supplementary skills. From the posterior probabilities of the target skills, we derived the BN-based CAT score, and the original CAT score from \cite{piatti_2022}, as defined in Section \ref{sec:casestudy}.
Fig.~\ref{fig:scatter} illustrates the correlation between these two scores for the following four models: the baseline model from \cite{softcom} (Model B), the baseline model with constraints (Model BC), the one which also includes the supplementary skills (Model BCS) and finally the enhanced model including both constraints and supplementary skills (Model ECS). 
In the scatterplot, the BN-based CAT score, originally in the $[0, 9]$ range, has been rescaled in the $[0, 4]$ range to obtain a more direct graphic comparison.
In all cases the Pearson correlation coefficient, indicated in Fig.~\ref{fig:scatter} as $\rho$, is very high, confirming consistency between the BN-based assessment and the one performed by experts.

\begin{table}[htb]
\setlength\extrarowheight{2pt}
\caption{
Inference times for all models.
\label{tab:times}}
\begin{center}
    \begin{tabular}{l|M{1.7cm}M{1.7cm}M{1.7cm}}
    \multicolumn{1}{c|}{{Model}} & Total inference time (sec) & Inference time per student (sec) & {Inference time per student per task (sec)}\\
    \midrule
    Model B    & 29.615  & 0.272 & 0.023 \\
    Model BC  & 28.940  & 0.266 & 0.022 \\
    Model BCS & 316.555 & 2.904 & 0.242\\
    Model ECS & 306.517 & 2.812 & 0.234\\                              
    \end{tabular}
\end{center}
\end{table}
\mbox{Table~\ref{tab:times}} shows the inference time of each model for all 12 tasks and all 109 students, as well as the estimated inference time to provide an assessment for a single student in all tasks and for a single student in a single task.
It is important to note that the inference times for a single student in a single task may vary depending on the complexity of the task. However, these average inference times provide a rough estimate of the performance of the four models in terms of inference time.
The total inference times of the four models can vary significantly, with Model B and Model BC being particularly fast at under 30 seconds, while Model BCS and Model ECS have a much longer inference time, taking over 5 minutes each. 
However, it's worth noting that when considering the assessment time for a single student, all models perform very quickly and can be considered suitable for real-time applications.

Besides the overall consistency of the summary metrics, the probabilistic assessments derived from the BN-based models provide more detailed information about student competence profiles in the form of posterior probabilities for each competence level and, when included, for the supplementary skills.

When looking at the posterior probabilities for the individual students, it is possible to understand their competence profile better and demonstrate the model interpretability. 
Moreover, by comparing competence profiles issued by the four models considered, it is possible to discern relevant differences between them and recognise the effects of the different improvements introduced in this work. 
To this goal, we summarise in Table~\ref{tab:answers} the answers provided by four representative pupils, in \mbox{Table~\ref{tab:scores}} the BN-based CAT scores versus the CAT score, and in Tables~\ref{tab:posteriors} and \ref{tab:posteriors_supp} the corresponding posterior probabilities inferred by the models for the target and the supplementary skills, respectively. 
The subset of students we use to demonstrate the performance of the models has been selected by choosing some pupils with interesting situations.

\begin{table}[htb]
\setlength\extrarowheight{2pt}
\caption{Comparison of the BN-based CAT scores with the CAT score from \mbox{\cite{piatti_2022}} of all models for a representative subset of pupils.\label{tab:scores}}
\begin{center}
\begin{tabular}{c|c|l}
{{Student}} & CAT score & BN-based CAT score  \\ 
\midrule
\multirow{4}{*}{{21}} &\multirow{4}{*}{3.3} &  2.23 (Model B) \\
                      &                     & 1.65 (Model BC)\\
                      &                     & 1.98 (Model BCS) \\
                      &                     & 2.00 (Model ECS)\\ \midrule
\multirow{4}{*}{{33}} &\multirow{4}{*}{0.75} & 2.00 (Model B) \\
                      &                      & 0.00 (Model BC)\\
                      &                      & 1.33 (Model BCS) \\
                      &                      & 1.47 (Model ECS)\\ \midrule
\multirow{4}{*}{{81}} &\multirow{4}{*}{1.75} & 2.90 (Model B) \\
                      &                      & 0.07 (Model BC)\\
                      &                      & 1.62 (Model BCS) \\
                      &                      & 1.82 (Model ECS)\\ \midrule
\multirow{4}{*}{{92}} &\multirow{4}{*}{2.5}  & 1.77 (Model B) \\
                      &                      & 1.42 (Model BC)\\
                      &                      & 1.59 (Model BCS) \\
                      &                      & 1.79 (Model ECS)\\
\end{tabular}
\end{center}
\end{table}

\begin{table*}[htb]
\centering
\caption{Answer to the 12 schemes, expressed as target and supplementary skills, for a representative subset of pupils.\label{tab:answers}}
\begin{tabular}{c|cccccccccccc}
{Student}& \multicolumn{1}{c}{{T1}} & \multicolumn{1}{c}{{T2}} & \multicolumn{1}{c}{{T3}} & \multicolumn{1}{c}{{T4}} & \multicolumn{1}{c}{{T5}} & \multicolumn{1}{c}{{T6}} & \multicolumn{1}{c}{{T7}} & \multicolumn{1}{c}{{T8}} & \multicolumn{1}{c}{{T9}} & \multicolumn{1}{c}{{T10}}& \multicolumn{1}{c}{{T11}}& \multicolumn{1}{c}{{T12}}\\
\midrule
\multirow{2}{*}{21}  &   1D-V	& 1D-V	  & 2D-V	 & 1D-V	     & 1D-V	     & 1D-V	   & 2D-V	& 2D-V	& 2D-V	  & 1D-V	& 1D-V	& 1D-V \\
&S$_{2}$	&S$_{2}$; S$_{6}$	&S$_{3}$; S$_{10}$	&S$_{3}$	&S$_{3}$; S$_{4}$	&S$_{6}$	&S$_{8}$; S$_{10}$	&S$_{1}$; S$_{5}$; S$_{10}$	&S$_{1}$; S$_{10}$	&S$_{1}$; S$_{4}$	&S$_{1}$	&S$_{1}$; S$_{5}$\\ \midrule
\multirow{2}{*}{33}  &   1D-V	& 1D-VS	  & 1D-VS	 & 1D-VSF	 & 1D-VS	 & 1D-VS   & 1D-VS	& \multicolumn{1}{c}{\multirow{2}{*}{fail}}	& \multicolumn{1}{c}{\multirow{2}{*}{fail}}	  & \multicolumn{1}{c}{\multirow{2}{*}{fail}}	& \multicolumn{1}{c}{\multirow{2}{*}{fail}}	& \multicolumn{1}{c}{\multirow{2}{*}{fail}} \\
                     &   S$_2$	& S$_2$; S$_{6}$	& S$_{3}$	& S$_{3}$	& S$_{3}$	 &  S$_1$; S$_{3}$ &  S$_{5}$	        & 	& & 	& 	&  \\\midrule
\multirow{2}{*}{81}  &   1D-V	& 1D-V	  & 1D-V	 & 1D-VS	 & 1D-V	     & 1D-V	   & 2D-VSF	& 0D-VS	& 2D-V	  & \multicolumn{1}{c}{\multirow{2}{*}{fail}}	& \multicolumn{1}{c}{\multirow{2}{*}{fail}}	& \multicolumn{1}{c}{\multirow{2}{*}{fail}} \\
                     &   S$_{2}$	& S$_{2}$; S$_{6}$	& S$_{3}$	& S$_{3}$	& S$_{3}$; S$_{4}$ & 	S$_{6}$	   &  S$_1$; S$_{5}$; S$_{10}$	& S$_{1}$	& S$_{1}$; S$_{10}$& 	& 	&  \\ \midrule
\multirow{2}{*}{92}  &   1D-V	& 1D-V	  & 1D-V	 & 1D-V	     & 1D-V	     & 1D-V	   & 0D-V	& 0D-V	& 0D-VSF  & 1D-VS	& 2D-V	& 0D-V \\
 &S$_{2}$ & S$_{2}$; S$_{6}$ & S$_{3}$ & S$_{3}$ & S$_{3}$; S$_{4}$ & S$_{6}$ & S$_{1}$ & S$_{1}$ & S$_{1}$ &S$_{4}$; S$_{5}$ & S$_{1}$; S$_{10}$ & 	S$_{1}$\\
\end{tabular}
\end{table*}

\begin{table*}[htb]
\setlength\extrarowheight{2pt}
\caption{Posterior probabilities \begin{math}P(X_{rc}=1|{Y})\end{math} of all models, for a representative subset of pupils.\label{tab:posteriors}}
\begin{center}
\begin{tabular}{c|l|ccccccccc}
\multirow{2}{*}{{Student}}& \multicolumn{1}{c|}{\multirow{2}{*}{{Model}}}& { \begin{math}X_{11} \end{math}}&{ \begin{math}X_{12} \end{math}}&{ \begin{math}X_{13} \end{math}}&{ \begin{math}X_{21} \end{math}}&{ \begin{math}X_{22}\end{math}}&{\begin{math}X_{23}\end{math}}&{ \begin{math}X_{31}\end{math}}&{ \begin{math}X_{32}\end{math}}&{ \begin{math}X_{33}\end{math}}\\ 
& & {0D-VSF}&{0D-VS}&{0D-V}&{1D-VSF}&{1D-VS}&{1D-V}&{2D-VSF}&{2D-VS}&{2D-V}\\ 
\midrule
\multirow{4}{*}{{21}} & Model B  & \gradient{0.50}	&	\gradient{0.51}	&	\gradient{0.67}	&	\gradient{0.51}	&	\gradient{0.57}	&	\gradient{0.96}	&	\gradient{0.59}	&	\gradient{0.83}	&	\gradient{0.80}\\
                              & Model BC & \gradient{1.00}	&	\gradient{1.00}	&	\gradient{1.00}	&	\gradient{1.00}	&	\gradient{1.00}	&	\gradient{1.00}	&	\gradient{0.69}	&	\gradient{0.38}	&	\gradient{0.07}\\
                              & Model BCS & \gradient{1.00}	&	\gradient{1.00}	&	\gradient{1.00}	&	\gradient{1.00}	&	\gradient{1.00}	&	\gradient{1.00}	&	\gradient{0.97}	&	\gradient{0.95}	&	\gradient{0.92}\\
                              & Model ECS & \gradient{1.00}	&	\gradient{1.00}	&	\gradient{1.00}	&	\gradient{1.00}	&	\gradient{1.00}	&	\gradient{1.00}	&	\gradient{1.00}	&	\gradient{1.00}	&	\gradient{1.00}\\ \midrule
\multirow{4}{*}{{33}} & Model B  & \gradient{0.00}	&	\gradient{0.00}	&	\gradient{0.00}	&	\gradient{0.00}	&	\gradient{1.00}	&	\gradient{0.00}	&	\gradient{0.00}	&	\gradient{0.00}	&	\gradient{0.00}\\
                              & Model BC & \gradient{0.00}	&	\gradient{0.00}	&	\gradient{0.00}	&	\gradient{0.00}	&	\gradient{0.00}	&	\gradient{0.00}	&	\gradient{0.00}	&	\gradient{0.00}	&	\gradient{0.00}\\
                              & Model BCS & \gradient{1.00}	&	\gradient{1.00}	&	\gradient{0.52}	&	\gradient{1.00}	&	\gradient{1.00}	&	\gradient{0.05}	&	\gradient{0.59}	&	\gradient{0.30}	&	\gradient{0.00}\\
                              & Model ECS & \gradient{1.00}	&	\gradient{1.00}	&	\gradient{0.69}	&	\gradient{1.00}	&	\gradient{1.00}	&	\gradient{0.39}	&	\gradient{0.63}	&	\gradient{0.33}	&	\gradient{0.03}\\ \midrule
\multirow{4}{*}{{81}} & Model B  & \gradient{0.03}	&	\gradient{0.00}	&	\gradient{0.00}	&	\gradient{0.00}	&	\gradient{0.00}	&	\gradient{1.00}	&	\gradient{0.00}	&	\gradient{0.00}	&	\gradient{0.00}\\
                              & Model BC & \gradient{0.01}	&	\gradient{0.00}	&	\gradient{0.00}	&	\gradient{0.00}	&	\gradient{0.00}	&	\gradient{0.00}	&	\gradient{0.00}	&	\gradient{0.00}	&	\gradient{0.00}\\
                              & Model BCS & \gradient{1.00}	&	\gradient{1.00}	&	\gradient{1.00}	&	\gradient{1.00}	&	\gradient{1.00}	&	\gradient{1.00}	&	\gradient{0.91}	&	\gradient{0.21}	&	\gradient{0.03}\\
                              & Model ECS & \gradient{1.00}	&	\gradient{1.00}	&	\gradient{1.00}	&	\gradient{1.00}	&	\gradient{1.00}	&	\gradient{1.00}	&	\gradient{0.95}	&	\gradient{0.67}	&	\gradient{0.44}\\ \midrule
\multirow{4}{*}{{92}} & Model B  & \gradient{0.55}	&	\gradient{0.40}	&	\gradient{0.41}	&	\gradient{0.33}	&	\gradient{0.13}	&	\gradient{1.00}	&	\gradient{0.46}	&	\gradient{0.05}	&	\gradient{0.00}\\
                              & Model BC & \gradient{1.00}	&	\gradient{0.99}	&	\gradient{0.99}	&	\gradient{0.76}	&	\gradient{0.70}	&	\gradient{0.68}	&	\gradient{0.13}	&	\gradient{0.00}	&	\gradient{0.00}\\
                              & Model BCS & \gradient{1.00}	&	\gradient{1.00}	&	\gradient{1.00}	&	\gradient{1.00}	&	\gradient{1.00}	&	\gradient{1.00}	&	\gradient{0.59}	&	\gradient{0.19}	&	\gradient{0.01}\\
                              & Model ECS & \gradient{1.00}	&	\gradient{1.00}	&	\gradient{1.00}	&	\gradient{1.00}	&	\gradient{1.00}	&	\gradient{1.00}	&	\gradient{0.79}	&	\gradient{0.58}	&	\gradient{0.41}\\                                 
\end{tabular}
\end{center}
\end{table*}

\begin{table*}[htb]
\setlength\extrarowheight{2pt}
\caption{Posterior probabilities \begin{math}P(S_{i}=1|{Y})\end{math} of Model BCS and Model ECS, for a representative subset of pupils.\label{tab:posteriors_supp}}
\begin{center}
\begin{tabular}{c|l|M{1cm}M{1cm}M{1cm}M{1cm}M{1cm}M{1cm}M{1cm}M{1cm}M{1cm}M{1cm}}
&& {S$_{1}$}&{S$_{2}$}&{S$_{3}$}&{S$_{4}$}&{S$_{5}$}&{S$_{6}$}&{S$_{7}$}&{S$_{8}$}&{S$_{9}$}&{S$_{10}$} \\ 
{Student}& \multicolumn{1}{c|}{{Model}} & paint dot & fill empty dots & rows columns & squares & diagonal & l & zigzag &  rows columns poly. & diagonals zigzag poly. & repetitions \\ 
\midrule
\multirow{2}{*}{{21}} & Model BCS & \gradient{1.00} &	\gradient{1.00}	& \gradient{1.00} &	\gradient{1.00} &	\gradient{1.00} &	\gradient{1.00} &	\gradient{0.40} &	\gradient{1.00} &	\gradient{0.26} &	\gradient{1.00}\\
                       & Model ECS & \gradient{1.00} &	\gradient{1.00}	& \gradient{1.00} &	\gradient{1.00} &	\gradient{1.00} &	\gradient{1.00} &	\gradient{0.52} &	\gradient{1.00} &	\gradient{0.38} &	\gradient{1.00}\\ \midrule
\multirow{2}{*}{{33}} & Model BCS & \gradient{1.00} & \gradient{1.00} & \gradient{1.00} & \gradient{0.42} & \gradient{1.00} & \gradient{1.00} & \gradient{0.38} & \gradient{0.15} & \gradient{0.16} & \gradient{0.13} \\
                      . & Model ECS & \gradient{1.00} & \gradient{1.00} & \gradient{1.00} & \gradient{0.43} & \gradient{1.00} & \gradient{1.00} & \gradient{0.42} & \gradient{0.21} & \gradient{0.22} & \gradient{0.19} \\ \midrule       
\multirow{2}{*}{{81}} & Model BCS & \gradient{1.00} & \gradient{1.00} & \gradient{1.00} & \gradient{1.00} & \gradient{1.00} & \gradient{1.00} & \gradient{0.36} & \gradient{0.34} & \gradient{0.31} & \gradient{1.00} \\
                       & Model ECS & \gradient{1.00} & \gradient{1.00} & \gradient{1.00} & \gradient{1.00} & \gradient{1.00} & \gradient{1.00} & \gradient{0.39} & \gradient{0.41} & \gradient{0.35} & \gradient{1.00} \\ \midrule
\multirow{2}{*}{{92}} & Model BCS & \gradient{1.00}	& \gradient{1.00}	& \gradient{1.00}	& \gradient{1.00}	& \gradient{1.00}	& \gradient{1.00}	& \gradient{0.34}	& \gradient{0.31}	& \gradient{0.31}	& \gradient{1.00}\\
                       & Model ECS & \gradient{1.00}	& \gradient{1.00}	& \gradient{1.00}	& \gradient{1.00}	& \gradient{1.00}	& \gradient{1.00}	& \gradient{0.36}	& \gradient{0.35}	& \gradient{0.35}	& \gradient{1.00}\\
\end{tabular}
\end{center}
\end{table*}

Pupil 21 is a high-performing student, in terms of CAT and BN-based scores,  who consistently used high-level artefacts for all CAT schemes and primarily used 2D algorithms.

Pupils 33 and 81 cannot be considered high-performing since they failed to complete some of the CAT schemes.
Pupil 33 solved only the first seven schemes, always using 1D algorithms and almost always relying on the VS artefact. In comparison, student 81 was successful in the first nine schemes where he applied different algorithms and artefacts, but mostly the 1D-V. 
For both students 33 and 81, the BN-based CAT scores predicted by the four models vary significantly, indicating that the models may be producing different predictions of their abilities. The difference between the original and BN-based CAT scores is inconsistent across the models. For both students, the largest difference between the original and BN-based CAT scores is observed in Model B, which predicts a much higher score for both students.
On the other hand, Model BC predicts a meager BN-based CAT score close to 0 for both students, indicating that this model may not be the most accurate for these particular students. This suggests that other models may be better suited for predicting their performance on the CAT.

Pupil 92's performance was strong, as he successfully completed all 12 tasks using different skill levels. 
He solved the first six schemes with the 1D-V skill, reduced the algorithm's complexity in the following ones, changed artefact for some of the more complex tasks, and applied the highest level skill, 2D-V, in a tricky schema.
Regarding the BN-based scores, all four models predicted a lower BN-based CAT score for student 92 than the original CAT score, although the differences were not as large as those observed for students 33 and 81. This suggests that student 92 is a relatively strong performer overall, but there is potential for improvement in his performance.

For all students, the baseline model assigns posterior probabilities equal or very close to one to the most used skill levels but fails to recognise that they also possess lower level skills, to which rather small probabilities, eventually equal to zero, as for the worst performing students 33 and 81, are assigned. 

On the one hand, when the ordering between skills is explicitly imposed, this problem is solved: the probabilities of lower skills increase, and those of higher skills decrease. 
This may lead to an excessive penalisation of higher skills, as in the case of pupils 33 and 81, where, as a consequence of the repeated failures in applying even the lowest competence level, Model BC decides for the total absence of the competencies under examination, returning a posterior probability of zero, even for the skills successfully used by the students in several schemes. These inferences look too severe for these situations,  where an expert would rather attribute the errors to the specific difficulties of the failed tasks rather than the total lack of algorithmic skills. 

On the other hand, when the supplementary skills are included in the assessment (Models BCS and ECS), this issue is solved, and the result of the posterior inference is consistent with the hierarchy of competencies defined by the rubric and the observations collected. 
In this case, the model understands that the failure follows from lacking the supplementary skills necessary to solve specific schemes with more complex algorithms and not from the lack of target skills.

For instance, according to models BCS and ECS, pupil 21 is likely to miss monochromatic zigzags (S$_7$) and polychromatic diagonals and zigzags (S$_9$), justifying the failure in applying the possessed 2D competence in schemes related to these supplementary skills.

Finally, employing more elaborate models, such as the ECS one, may, in some cases, reward the ability to apply high-level skills in more complex tasks, i.e., those assigned with higher inhibition probabilities, such as for pupil 92 who managed to solve scheme T11, a difficult one according to the parameters' elicitation in Fig.~\ref{fig:its}, using a 2D-V skill and thus 2D algorithms are given a much higher probability by Model ECS than by Model BCS.

\subsection{Discussion}

Recent literature shows that BNs are among the most common tools for modelling student knowledge in ITS \mbox{\cite{mousavinasab2021intelligent}}, however, in most works, a great deal of effort is devoted to building an ad-hoc network and eliciting its parameters \mbox{\cite{RAMIREZNORIEGA20171488,eryilmaz2020development}} or collect data from which they could be learned \mbox{\cite{hooshyar2018sits,wu2020student,rodriguez2021bayesian}}. 
This effort and technical complexity might be one reason why most ITSs in the literature concern primarily technical domains. For instance, \mbox{\cite{soofi2019systematic}} shows that most ITSs focus on computer science education. If the reason for this phenomenon is that experts in this field are more comfortable using mathematical models, our proposal to base BN model elicitation on assessment rubrics, a popular educational tool for eliciting the relation between skills and behaviours, opens up the possibility of bringing this method closer to other, less technical disciplines.

Our study estimated the performance of four different models (Model B, Model BC, Model BCS and Model ECS) in assessing the algorithmic competence of 109 pupils based on their answers to 12 tasks.
The models were evaluated by comparing the BN-based CAT score derived from the posterior probabilities of the target skills with the original CAT score obtained from the expert assessment. 
The results show a high correlation between the two scores in all models, indicating consistency between the BN-based assessment and the one performed by experts.
The probabilistic assessments derived from the BN-based models provided more detailed information about student competence profiles in the form of posterior probabilities for each competence level and supplementary skills. 
The analysis of the models' posterior probabilities for the individual students highlighted relevant differences between the four models showing the effects of the improvements introduced in the study.
The model interpretability was demonstrated by comparing competence profiles issued by the four models considered. 
The study found that Model B assigned posterior probabilities close to one of the most used skill levels but failed to recognise that the students also possessed lower-level skills, thus violating the ordering between competencies assumed by the assessment rubric. 
Instead, this issue was solved when the ordering between skills was explicitly imposed (Model BC). 
The result of the posterior inference was consistent with the hierarchy of competencies defined by the rubric. 
Finally, including the supplementary skills in the assessment (Model BCS) and refining the model's parameters (Model ECS) produced more detailed competence profiles and the evaluation quality more consistent with the observations collected and their interpretation by the experts.

Our study has demonstrated the effectiveness and efficiency of our approach to student knowledge modelling. Specifically, we found that the noisy-OR network structure can model student knowledge with high accuracy and minimal effort once the assessment rubric is defined.
We first evaluated the baseline model, defined by only two parameters (i.e., \textit{slip} and \textit{guess} probabilities) assumed to be constant across questions and skills. 
Even with this minimal parameterisation, we found that the inferences about the learner's knowledge were consistent with the expert assessment. 
This suggests that the model is effective even when the number of parameters is small and can serve as a starting point for subsequent refinements.
We have illustrated this refinement process in the enhanced model, which used different values of the inhibition parameters to describe the different task complexities while keeping the elicitation cost limited thanks to some simplifying assumptions. This refinement was shown to improve the quality of the assessment. 
Further refinements could be introduced by removing some of the assumptions.

Our approach is highly practical and efficient for real-world educational settings. The initial effort required for elicitation is almost equivalent to that used to define the rubric, and subsequent refinements can be made with little additional exertion. Findings suggest that our approach can facilitate accurate and efficient student knowledge modelling for various educational contexts and teachers with a non-technical background.

\section{Conclusion}\label{sec:concl}
This article is an extension of the work initially presented in the SoftCOM conference paper \cite{softcom}, which proposed a procedure for deriving a learner model for automatic skill assessment directly from the competence rubric of any set of tasks. 
This study aims to solve the two main limitations of the previous approach.
Firstly, although partially implied by the structure of the model and implicitly defined in the assessment rubric, the ordering between competence levels was not strictly enforced in \cite{softcom}.
Secondly, in the previous modelling approach, either disjunctive or conjunctive gates must be used, while sometimes it would be helpful to combine both or implement more general relations between competencies. Supplementary skills, complementary to those in the rubric and not under assessment but necessary for the accomplishment of the task, were not included in the model.
In this article, we provide a method to fulfil both needs.
The first one is met by introducing dummy observed nodes in the network without changing its structure.
The second was achieved by designing a network with two layers of gates, the first performing the disjunctive operation through noisy-OR gates and the second executing the conjunctive one by a logical AND. 

The approach has been illustrated, and its feasibility demonstrated by implementing the automatic assessment of the Cross Array Task (CAT) \cite{piatti_2022}. 
Four models based on noisy gates Bayesian networks (BNs) have been developed and evaluated. By comparing the BN-based CAT score obtained for the 109 pupils performing the CAT to the original scores assigned by the experts, we have shown the consistency of the four models developed with the expert assessment. 
Moreover, by analysing the more detailed individual student assessment provided by the posterior probabilities, we have shown that by adding constraints and supplementary skills to the original baseline model, as well as by refining its parameters, it was possible to improve the quality of the assessment, its consistency the student behaviours, and the coherence between the model outcomes and the partial ordering between competencies defined by the assessment rubric, increasing the modelling tool's flexibility and allowing for a more accurate result.
Outcomes also confirmed that changes from our previous modelling approach do not compromise the model compact parametrisation, its simple elicitation by experts and its interpretability, derived from using noisy gates BNs \cite{Anonymous2022}. 

The proposed approach offers a promising practical, and efficient student knowledge modelling solution. By allowing for a straightforward translation of assessment rubrics into a flexible and scalable mathematical model, this approach enables more widespread use of Intelligent Tutoring and Assessment Systems (ITAS) that can interact with learners automatically and in real time. The combination of assessment rubrics and the noisy-OR network structure makes this approach useful for teachers seeking to enhance student learning and performance, especially those without a technical background.

While the presented work focuses mainly on expert-based elicitation, ongoing work explores the possibility of improving the model by updating its parameters based on the evidence collected during user interaction.


%




\ifCLASSOPTIONcaptionsoff
  \newpage
\fi



\bibliographystyle{IEEEtran}
\bibliography{biblio}
%



\newpage
%
\begin{IEEEbiography}[{\includegraphics[width=1in,height=1in,clip,keepaspectratio]{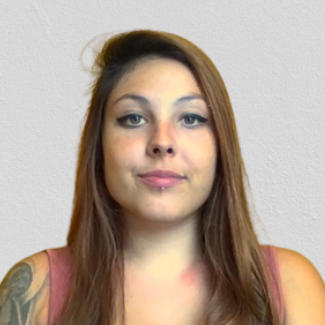}}]{Giorgia Adorni} is a PhD candidate at the Swiss AI Lab IDSIA, joint institute USI-SUPSI, with a background in Computer Science and Informatics. Her research interests include Artificial Intelligence, Machine Learning, Computer Science Education, Educational Robotics, Learning Technologies, and Human-Robot Interaction.
\end{IEEEbiography}
\begin{IEEEbiography}[{\includegraphics[width=1in,height=1in,clip,keepaspectratio]{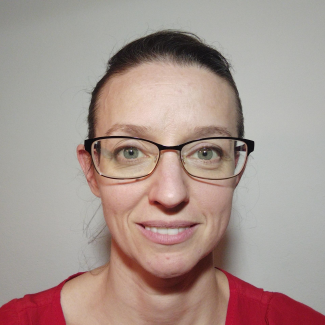}}]{Dr. Francesca Mangili} is a Senior Researcher at IDSIA, the Swiss AI Lab affiliated with the two Universities USI and SUPSI. She teaches courses on probability and applied statistics at SUPSI and has authored more than 40 peer-reviewed publications. Her main research interests are in the area of statistics, probabilistic reasoning and machine learning with applications in various fields among which the medical and educational ones stand out.
\end{IEEEbiography}
\begin{IEEEbiography}[{\includegraphics[width=1in,height=1in,clip,keepaspectratio]{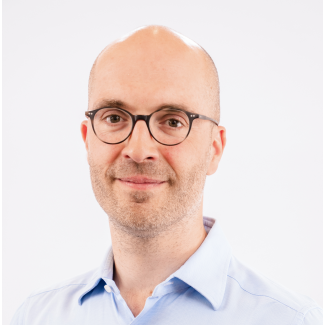}}]{Alberto Piatti} is director of the Department Education and Learning of the University of Applied Sciences and Arts of Southern Switzerland and professor in STEM Education. He has studied mathematics at the Swiss Federal Institute of Technology in Zürich and holds a PhD in Economics from the University of Lugano. 
\end{IEEEbiography}
\begin{IEEEbiography}[{\includegraphics[width=1in,height=1in,clip,keepaspectratio]{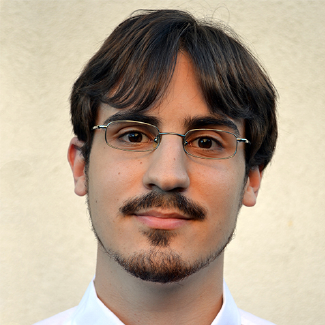}}]{Claudio Bonesana} is a Machine Learning Engineer and Researcher at IDSIA, the Swiss AI Lab affiliated with the two Universities USI and SUPSI. His main research interests are in the area of machine learning, distributed systems, and graphical models for AI and their application to real world problems.
\end{IEEEbiography}
\begin{IEEEbiography}[{\includegraphics[width=1in,height=1in,clip,keepaspectratio]{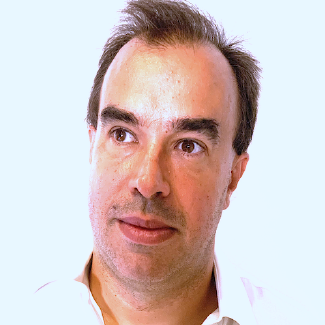}}]{Dr. Alessandro Antonucci} is a Senior Lecturer-Researcher at IDSIA, the Swiss AI Lab located in Lugano affiliated with the two Universities of Southern Switzerland. He is the author of more than 100 peer-reviewed publications and an Area Editor of the International Journal of Approximate Reasoning. His research interests mostly focus on probabilistic graphical models for AI reasoning and machine learning and their application to explainability and causality theory. 
\end{IEEEbiography}
\hfill






\end{document}